\documentclass[10pt,twocolumn,letterpaper]{article}

\usepackage[final,applications]{wacv} 
\usepackage{times}
\usepackage{epsfig}
\usepackage{graphicx}
\usepackage{amsmath}
\usepackage{amssymb}
\usepackage{booktabs}
\usepackage{adjustbox}
\usepackage{multirow}
\usepackage[table,xcdraw,dvipsnames]{xcolor}
\usepackage{color, colortbl}
\usepackage{nicefrac}
\usepackage[accsupp]{axessibility}

\definecolor{Gray}{gray}{0.9}
\definecolor{brightturquoise}{rgb}{0.85, 1, 1}
\definecolor{newpurple}{HTML}{BC61F5}%EF9B0F
\newcommand{\mycc}{\cellcolor[HTML]{EDF6FF}}
\newcommand{\myccnew}{\cellcolor[rgb]{0.61, 0.87, 1.0}}
\newcommand{\reffig}[1]{\text{Figure~\ref{#1}}}
\newcommand{\reftab}[1]{\text{Table~\ref{#1}}}
\newcommand{\refeq}[1]{\text{Eq.~\ref{#1}}}
\newcommand*{\affaddr}[1]{#1}
\newcommand*{\affmark}[1][*]{\textsuperscript{#1}}

\newcommand\blfootnote[1]{%
  \begingroup
  \renewcommand\thefootnote{}\footnote{#1}%
  \addtocounter{footnote}{-1}%
  \endgroup
}

\usepackage[pagebackref=true,breaklinks=true,colorlinks,bookmarks=false,citecolor=blue,linkcolor=red]{hyperref}
\begin{document}

%%%%%%%%% TITLE
\title{Do VSR Models Generalize Beyond LRS3?}

\author{Yasser Abdelaziz Dahou Djilali\affmark[1,2] \quad 
Sanath Narayan\affmark[1] \quad 
Eustache Le Bihan\affmark[1] \vspace{0.1cm}\\ 
Haithem Boussaid\affmark[1] \quad
Ebtessam Almazrouei\affmark[1] \quad
Merouane Debbah\affmark[1] \vspace{0.1cm}\\
\affaddr{\affmark[1]Technology Innovation Institute, UAE} \quad 
\affaddr{\affmark[2]Dublin City University, Ireland}
}

\maketitle
% \thispagestyle{empty}

%%%%%%%%% ABSTRACT
\begin{abstract}

The Lip Reading Sentences-3 (LRS3) benchmark has primarily been the focus of intense research in visual speech recognition (VSR) during the last few years. As a result, there is an increased risk of overfitting to its excessively used test set, which is only one hour duration. To alleviate this issue, we build a new VSR test set named WildVSR, by closely following the LRS3 dataset creation processes. We then evaluate and analyse the extent to which the current VSR models generalize to the new test data. We evaluate a broad range of publicly available VSR models and find significant drops in performance on our test set, compared to their corresponding LRS3 results. Our results suggest that the increase in word error rates is caused by the models’ inability to generalize to slightly “harder” and in the wild lip sequences than those found in the LRS3 test set. Our new test benchmark is made public in order to enable future research towards more robust VSR models.

\blfootnote{Code: \url{https://github.com/YasserdahouML/VSR_test_set}}

\end{abstract}

%%%%%%%%% BODY TEXT
\section{Introduction}

The primary objective of machine learning revolves around training models that are capable of better generalization. Typically, the quantification of generalization occurs through evaluating a model's performance on a held-out test set. The question then arises: what does satisfactory performance on this test set indicate? At the very least, it is desirable that the model exhibits similar performance on a new test set derived from the same data creation process. In this work, we study these questions for the problem of Visual Speech Recognition (VSR).

Indeed, most perception problems are interpolative in their nature~\cite{chollet2019measure} and satisfy the manifold hypothesis~\cite{fefferman2016testing}. These tasks are intuitive for humans, and are usually solved in the early layers of the visual cortex in a matter of milliseconds (\ie, classification, recognition, \etc) \cite{kosaka2003neural, todorov2012role}. For such problems, deep learning is a perfect fit with its ability to perform non-linear interpolation in a complex high-dimensional manifold, enabling arbitrary complex behavior~\cite{summerfield2022natural,chollet2019measure}, allowing for better generalization. However, lip-reading experts allude to a high-level step-wise and iterative reasoning to solve the task. This likely suggests that VSR has some higher level of extrapolation as compared to the common perception tasks. Thus, extensive model's generalization study is highly needed for the field. However, the one hour LRS3 test set is the main focus for evaluation state-of-the-art models.

In this paper, we follow the LRS3 \cite{afouras2018lrs3} creation process to build a new set from the wild. As expected, we observe that all VSR SoTA models fail to reach their reported Word Error Rate (WER) from LRS3 on the new test set. Nevertheless, the WER drops by an average of 30 points, our experiments reveal that the comparative ranking of models remains remarkably consistent when evaluated on our fresh test sets. Specifically, the models that exhibit the highest accuracy on the original test sets also demonstrate the highest accuracy on the new test sets. This suggests that the WER drops are not a result of  extensive hyper-parameters tuning that fit the particular lip sequences found in the initial test set. We study why this phenomenon arises, specifically, with the following contributions:

\begin{itemize}
    \item We present a new VSR test set, WildVSR, incorporating higher visual diversity, and spoken vocabulary.
    \item We benchmark existing models on the new test set, and find a clear performance drop.
    \item Nevertheless, we show a diminishing return in performance \vs compute, where self-supervised approaches consume significantly higher training compute budget for a moderate performance. 
    \item We propose a  new metric that accounts for a model's confidence, which improves the WER based ranking.
\end{itemize}

\section{Related Work}
% Here, we briefly discuss the works related to the task of visual speech recognition.

\noindent\textbf{State-of-the-art approaches:}
The work of \cite{ma2021end} proposed a curriculum learning approach, where shorter sequences are initially used for training followed by progressively adding longer ones. Differently, VTP \cite{prajwal2022sub} proposed sub-words learning scheme using frame-word boundaries to crop out training samples for a better convergence. These training strategies are computationally demanding and hard to scale to larger datasets. The recent works of \cite{ma2022visual, Afouras2019ASRReading} proposed to exploit the audio latent representations as part of a auxiliary task, where the latent features from the encoder are optimized to predict pretrained ASR representations, in addition to decoding the target text. This extra supervision through the ASR representations makes the optimization more stable and improves the convergence. Intuitively, if the transformer encoder is able to match the audio features statistics in earlier layers, it is easier to adjust the attention weights in the later layers for improved text decoding. 

% \noindent\textbf{Self-supervised models:} 
Another line of research leverages cross-modal pretraining on large datasets in a self-supervised way (SSL), followed by finetuning on labeled  video-text pairs \cite{shi2022learning,shi2022robust,haliassos2022jointly,zhu2022vatlm, ma2021lira}. The work of AV-HuBERT \cite{shi2022learning} fuses the masked audio-visual representations to predict the cluster assignments created from the audio features, thus, distilling knowledge from the audio stream features to model visual inputs. VATLM \cite{zhu2022vatlm} extends this by attempting to unify the modalities using a one tower design, where a single network is optimized to construct a common representation space for video, audio and text. This is achieved by setting a unified tokenizer for all modalities, and then performing the masked prediction task over the unified tokens. The works of \cite{sheng2021cross,Ma2021ContrastiveRepresentations} designed cross-modal self-supervised learning frameworks by adopting contrastive learning \cite{jaiswal2020survey} to learn discriminative visual representations that appear to improve VSR performance and generalization. RAVen \cite{haliassos2022jointly} designed an asymmetric SSL framework to induce a one-way knowledge distillation, where the audio networks predict both audio and video representations, whereas the visual network is restricted to predict the audio features only. This forces the audio network to serve as a strong teacher, as it would adjust to both modalities at the same time. 

More recently, Auto-AVSR~\cite{ma2023auto} proposed to obtain text transcripts for unlabelled datasets using pre-trained ASR models. These auto-labeled video-text pairs along with manually labeled datasets were then utilized to train conformer-based AVSR models in a fully-supervised manner. Furthermore, SynthVSR~\cite{liu2023synthvsr} first learned a speech-driven lip animation model on unlabeled audio-visual datasets and generated a synthetic dataset, which was utilized along with labeled and auto-labeled datasets, similar to~\cite{ma2023auto}, for training the VSR model.

\noindent\textbf{Datasets:} 
Typically, most VSR approaches exploit the popular LRS3~\cite{afouras2018lrs3} dataset for training the models. While the use of LRW~\cite{chung2017lip} and LRS2~\cite{son2017lip} is lesser due to their restrictive licences, different strategies are commonly employed to increase the training data. \Eg, using unlabeled/machine-labeled (AVSpeech~\cite{ephrat2018looking}, VoxCeleb2~\cite{chung2018voxceleb2}), generating synthetic videos (3.6k hours in \cite{liu2023synthvsr}), collecting large-scale non-public datasets (YT31k~\cite{makino2019recurrent}, YT90k~\cite{serdyuk2021audio}), \etc. However, for evaluating the VSR models, the aforementioned works extensively utilize the publicly-available LRS3 test set of one hour duration, thereby increasing the risk of models overfitting to LRS3 test set distribution. While the works of~\cite{makino2019recurrent,serdyuk2021audio,chang2023conformers} additionally evaluate on YTDEV18~\cite{makino2019recurrent} or MEET360~\cite{chang2023conformers}, these test benchmarks are private. In this work, we set out to gather a challenging evaluation benchmark for VSR and analyse the performance of available models in-depth. Our proposed benchmark will be made public for aiding future research in VSR to build more robust models that generalize better to  videos in the wild.

% \section{Evaluation Setup}
\section{Building the WildVSR Test Set}

Utilizing YouTube as a data source has become a popular approach for constructing audio and/or visual speech recognition datasets since this platform offers access to a vast amount of audio-visual content. However, developing a quality VSR test set requires careful processing of raw videos to create pairs of visual utterances, namely visual lip movement, and transcriptions. Given the computational cost associated with this procedure, it is crucial to meticulously filter and extract relevant content from the vast YouTube repository before processing it. Thus, the overall approach should consist of two stages: 1) select relevant YouTube videos, and 2) process the selected videos to construct the aforementioned pairs. Regarding the first stage, previous works have employed two different approaches.

First, in the work by Makino \etal~\cite{makino2019recurrent}, there is no mention of their video selection method. They extract snippets from YouTube videos where audio and transcripts match and then process these snippets to extract face tracks that align with the transcript. However, this approach does not guarantee that the initially extracted snippets from YouTube videos will display lip movements at all, as no strategy to select those videos is used. As a result, all extracted snippets have to be processed in search of face tracks while they may not be present at all, \eg, a screen recorded tutorial might not have a face track although the audio matches with the corresponding transcript. This can impose a significant computational burden. Still, they managed to build a 25 hours test set of 20k utterances called YTDEV18 but it is worth highlighting that this test set is not publicly available.

A second approach \cite{afouras2018lrs3}, employed to create the publicly available and thus widely used LRS3 dataset, involves restricting the content sources to high-quality videos, focusing on TED and TEDx talks. These talks have reliable transcripts and visuals very likely correspond to what is targeted, maximizing the input-output rate of a processing pipeline intended to build a VSR dataset. However, this approach drastically reduces the available content from YouTube and may limit the dataset's ability to represent “in the wild” performances, as TED talks are often filmed in similar visual context, with underrepresented individuals and a formal language register.

Our approach tends to blend both strategies by profiting from the diversity of available data on YouTube as in ~\cite{makino2019recurrent} while targeting quality content likely to match our requirements as in \cite{afouras2018lrs3}. It also has the advantage of being scalable and language-oriented, meaning that the aim is to be computationally efficient and usable to target languages other than English. Moreover, as done in \cite{afouras2018lrs3}, it frees from legal issues encountered with former public datasets LRW~\cite{chung2017lip} and LRS2~\cite{son2017lip}, by targeting only free-to-use content.

\subsection{Data Gathering}
We leverage the power of YouTube search API that offers multiple filters solutions:
(\textit{i}) Search using keywords, (\textit{ii}) target most relevant videos in regard of a specified language (here English) and (\textit{iii}) target only videos published under creative commons license.

We start from a list of $21$ keywords, including interview or discussion for example,  in order to get the most relevant content, expanding the content-oriented filtering explored in \cite{afouras2018lrs3}. Furthermore, we ensure that these videos do not overlap with LRS3 by cross-verifying the YouTube ids. The resulting videos are then processed as described next.

\subsection{Data Processing}
Each video is divided into multiple shots using a scene change detection~\cite{7040826} method based on changes in three-dimensional histograms and faces in each frame of the video are detected using the YOLOv5n0.5-Face~\cite{qi2023yolo5face} for its unmatched precision-computation cost ratio. The detected faces are then matched and tracked across the frames to obtain multiple face tracks. Afterwards, we utilize SyncNet~\cite{chung2017out} for filtering these tracks through active speaker detection and obtain face track segments with active speakers corresponding to the audio of the track. We employ the ASR model Whisper ~\cite{radford2022robust} to detect language for discarding non-English tracks and to obtain pseudo-transcripts. Using the word timestamps provided in Whisper results, tracks are then segmented into clips with durations ranging from $0.5$ to $16$ seconds. These clips, along with their transcripts, are further verified manually to ensure high-quality clip-text pairs spanning a total of $4.8$ hours in duration, and a total of $2854$ utterances coming from $478$ YouTube videos.

\subsection{Test Set Quality Evaluation}
\begin{figure}[t]
  \centering
  \includegraphics[width=\columnwidth]{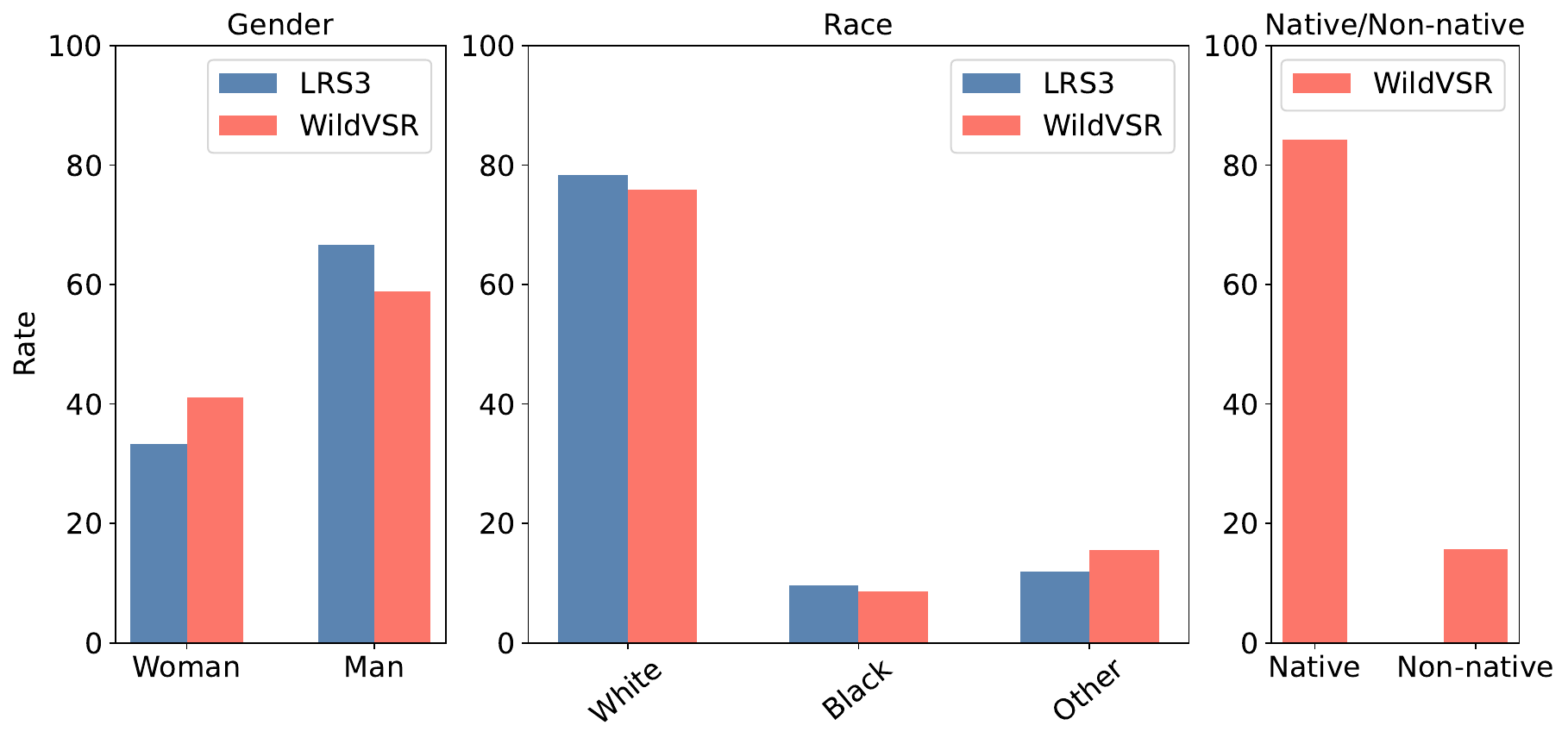}\vspace{-0.2cm}
  \caption{\textbf{Comparison between LRS3 and proposed test sets in terms of gender and race attributes.} Compared to the LRS3 test set, we observe a marginal improvement for race attribute (in the center) and relatively better diversity in terms of gender (on the left) for our test set.\vspace{-0.3cm}}
  \label{fig:gender_race}
\end{figure}

Vocabulary size and visual content variability are two aspects to consider when evaluating a VSR test set quality. As shown in \reftab{tab:stats_compare}, our WildVSR test set achieves a clear increase in vocabulary size, encompassing 71\% of the vocabulary from LRS3, along with an increase in number of unique speakers, utterances, word instances, and total duration. We use VGG-Face~\cite{serengil2020lightface} to identify the different speakers present across the test set. Afterward, we manually review and verify all speakers, resulting in $618$ distinct identities distributed across $478$ YouTube videos. Furthermore, we employ Deepface framework~\cite{serengil2021lightface} to obtain coarse-level demographic attribute metrics, namely gender and race. These attributes are then verified manually. Additionally, we rely on subjective analysis to determine whether a speaker's accent is native or not based solely on fluency. Compared to LRS3 test set Deepface predictions (not verified manually) as shown in \reffig{fig:gender_race}, our proposed test set demonstrates a slightly more balanced distribution across all attributes, indicating improved diversity. However, it is important to note that biases inherent to the online platform may be present in the dataset.

\begin{table}[t] 

    \centering
    \caption{\textbf{Comparison of statistics between LRS3 and our proposed test sets}. Our proposed test set has higher number of utterances along with $1.5\times$ unique speakers, $4.6\times$ word instances, $3\times$ vocabulary coverage and $5.3\times$ the duration.\vspace{-0.2cm}}
    \label{tab:stats_compare}

    \setlength{\tabcolsep}{8pt}
    \adjustbox{width=1\columnwidth}{
    \begin{tabular}{l| c |c| c |c |c}  
        \toprule[0.1em]
        
            \textbf{Test set}  & \textbf{\# Spk}. & \textbf{\# Utt.} & \textbf{Word inst.} & \textbf{ Vocab }& \textbf{Hours }  \\
            \midrule
            LRS3      &  412     & 1,321  & 9,890 & 1,997 & 0.9 \\
          \textbf{WildVSR}      & \textbf{618}        &  \textbf{2,854} & \textbf{45,182} & \textbf{6,040} & \textbf{4.8}      \\
            
        \bottomrule[0.1em]
    \end{tabular}
    \vspace{-0.7cm}
    }
\end{table}

\section{Experiments}

\begin{table*}[t] 
    \centering
    \setlength{\tabcolsep}{9pt}

    \caption{\textbf{Performance comparison on our proposed test set in low-resource and high-resource settings.} The performance is reported in terms of WER and proposed $Rank_{wer}$ metrics. `Base' and `Large' denote the size of the self-supervised video encoder employed. Performance of supervised approaches are also reported. The LRS3 test set performance is also shown for reference. We also report the performance of ASR (audio only) and AVSR (audio-visual) models on both LRS3 and our WildVSR test sets, in addition to the compute budget required (in ExaFLOPs) for training the respective AVSR and VSR models. \vspace{-0.2cm}}
    % \vspace{2mm}

    % \begin{threepartable}
    
     \adjustbox{width=\textwidth}{
    \begin{tabular}
    {c l c c c c  c c | c  c} 
     \toprule[0.1em]
      & \textbf{Method} & \textbf{Unlabeled}  & \textbf{Labeled} & \textbf{Decoding} & \textbf{Compute} & \multicolumn{2}{c|}{\textbf{$WER$}} & \multicolumn{2}{c}{\textbf{$Rank_{wer}$}} 
    \\
      \cmidrule{7-10}
     & & \textbf{AV data} & \textbf{data} & & (ExaFLOPs) &\textbf{LRS3}  & \mycc \textbf{WildVSR} & \textbf{LRS3}  & \mycc \textbf{WildVSR}
      \\
     \toprule[0.1em]

     \multirow{2}{*}{\textbf{ASR Models}} 

    & Wav2vec2.0 \cite{baevski2020wav2vec} & -- & -- & CTC & -- &6.1 & \mycc 17.7 & 6.9 & \mycc 18.0  \\
    & Whisper \cite{radford2022robust} & -- & -- & CE & -- &1.1 & \mycc 4.2  & 1.3 & \mycc 4.8 \\
    \midrule
    \midrule

    \multirow{4}{*}{\textbf{AVSR Models}} 

    & Base AV-HuBERT~\cite{shi2022learning}& 1759h & 30h &   CE & 39.7  & 4.1 &  \mycc 24.4    & 4.3  &\mycc 26.1  \\
    & Base AV-HuBERT~\cite{shi2022learning}& 1759h & 433h &   CE & 39.9 &  1.8 & \mycc 13.7    &  2.1 &\mycc  14.2 \\
    & Large AV-HuBERT~\cite{shi2022learning}& 1759h & 30h &   CE & 106.7 & 3.4 &  \mycc 22.8   &  3.6 &\mycc  24.3 \\
    & Large AV-HuBERT~\cite{shi2022learning}& 1759h & 433h &   CE & 107.3 & 1.5 &  \mycc 12.9  &  1.7 &\mycc  13.2  \\

    \midrule
    \midrule
    
    \multicolumn{10}{c}{\myccnew\textit{Low-resource setting}} \\
    \multirow{4}{*}{\textbf{Self-Supervised}}  & 

    AV-HuBERT~\cite{shi2022learning}& 433h & 30h &   CE & 39.7 &  51.8  &  \mycc 79.4 & 67.0 &\mycc  83.3 \\
    
     \multirow{4}{*}{\textbf{Base}} & RAVen~\cite{haliassos2022jointly} & 433h & 30h &  CTC+CE & 3.7 &  48.1 & \mycc 75.4 & 64.2 & \mycc 80.4 \\

     & AV-HuBERT~\cite{shi2022learning}& 1759h & 30h &  CE & 39.7 &  46.1 & \mycc 73.2 & 62.2 & \mycc 77.5 \\
     & RAVen~\cite{haliassos2022jointly} & 1759h & 30h &  CTC+CE & 14.9 &  40.2 & \mycc 66.9 & 54.1 & \mycc 72.1  \\
    
     \midrule
     \multirow{4}{*}{\textbf{Self-Supervised}}  & AV-HuBERT~\cite{shi2022learning}& 433h & 30h &  CE & 106.7 &  44.8 & \mycc 75.7 & 59.5  & \mycc 81.8 \\% 

     \multirow{4}{*}{\textbf{Large}} & AV-HuBERT~\cite{shi2022learning}& 1759h & 30h &  CE & 106.7 &  32.5 & \mycc 61.9 & 41.1 & \mycc 68.0 \\
     & AV-HuBERT~\cite{shi2022learning} w/ self-training& 1759h & 30h &  CE & 106.7 &  28.6 & \mycc 52.4 & 37.7 & \mycc 58.6 \\
     & RAVen~\cite{haliassos2022jointly} & 1759h & 30h &   CTC+CE & 96.8 &  32.5 &   \mycc 57.7 &  41.6 & \mycc 63.4\\
      & RAVen~\cite{haliassos2022jointly} w/ self-training & 1759h & 30h &   CTC+CE & 131.7 &  23.8 &  \mycc 48.4 & 31.5  & \mycc 52.0 \\

    \midrule
    \midrule
    \multicolumn{10}{c}{\myccnew\textit{High-resource setting}} \\
    
    \multirow{6}{*}{\textbf{Supervised}} & Ma \etal \cite{ma2022visual}& -- & 1459h & CTC+CE & 2.1 & 32.3 & \mycc 58.4 & 42.4 & \mycc 63.6 \\
    & Prajwal \etal \cite{prajwal2022sub} & -- & 698h & CE & $\dagger$ & 40.6 & \mycc 75.6 & 57.3 & \mycc 86.9 \\
    & Prajwal \etal \cite{prajwal2022sub} & -- & 2676h & CE & $\dagger$ & 30.7 & \mycc 68.7 & 43.7 & \mycc  82.1 \\

    & Auto-AVSR~\cite{ma2023auto} & -- & 661h & CTC+CE & 6.7 & 32.7 & \mycc 62.3  & 42.5  & \mycc 67.1 \\
    & Auto-AVSR~\cite{ma2023auto} & -- & 1759h & CTC+CE & 17.8 & 25.1 & \mycc 49.3  & 31.7 & \mycc 53.2 \\
    & Auto-AVSR~\cite{ma2023auto} & -- & 3448h & CTC+CE & 34.9 & \textbf{19.1} & \mycc \textbf{38.6} & \textbf{24.8} & \mycc \textbf{41.8} \\

    \midrule
      \multirow{4}{*}{\textbf{Self-Supervised}} & AV-HuBERT~\cite{shi2022learning}& 433h & 433h & CE & 39.9 & 44.0 &  \mycc 71.6  & 58.1 & \mycc 76.2 \\ %\tnote{2}\\
     \multicolumn{1}{c}{\multirow{4}{*}{\textbf{Base}}} & RAVen~\cite{haliassos2022jointly} & 433h & 433h & CTC+CE & 5.0 & 39.1 & \mycc 69.9 & 52.8 & \mycc 77.6 \\

     & AV-HuBERT~\cite{shi2022learning}& 1759h & 433h& CE & 39.9 & 34.8 & \mycc 58.1 & 50.5 & \mycc 63.9 \\
     & RAVen~\cite{haliassos2022jointly} & 1759h & 433h & CTC+CE & 16.3 & 33.1 & \mycc 60.0 & 42.6 & \mycc 65.7 \\
     
     \midrule

     \multirow{4}{*}{\textbf{Self-Supervised}}  & AV-HuBERT~\cite{shi2022learning}& 433h & 433h & CE & 107.3 & 41.6 & \mycc 69.4 & 56.1 & \mycc 73.5 \\% \tnote{2}\\

     \multicolumn{1}{c}{\multirow{4}{*}{\textbf{Large}}} & AV-HuBERT~\cite{shi2022learning}& 1759h & 433h & CE & 107.3 & 28.6 & \mycc 51.7 & 37.4 & \mycc 55.9 \\
     & AV-HuBERT~\cite{shi2022learning} w/ self-training & 1759h & 433h & CE & 107.3 & 26.9 & \mycc 48.7 & 34.9 & \mycc 53.0\\
     & RAVen~\cite{haliassos2022jointly} & 1759h & 433h& CTC+CE & 105.3 & 27.8 & \mycc 52.2 & 36.6  & \mycc 55.3\\
     & RAVen~\cite{haliassos2022jointly} w/ self-training  & 1759h & 433h& CTC+CE & 131.7 & 23.1 & \mycc 46.7 & 30.8 & \mycc 49.8 \\

     \bottomrule[0.1em]
    \end{tabular}}
    \label{sota_high_res}
\end{table*}

Here, we evaluate a broad range of VSR models spanning five years of progress in a highly active area of research. The models include the fully-supervised models: Ma \etal~\cite{ma2022visual}, VTP~\cite{prajwal2022sub}, Auto-AVSR~\cite{ma2023auto} and a set of self-supervised models fine-tuned on LRS3 with the different pretraining regimes such as RAVen~\cite{haliassos2022jointly} and AV-HuBERT~\cite{shi2022learning}. The VSR models generally comprise a ResNet-3D frontend followed by a transformer encoder-decoder. The frontend encodes the video lip sequence into a temporal sequence of features, which is fed into the encoder. The decoder then autoregressively decodes the text tokens by cross-attending to the encoder output features. Among the above models, Ma \etal and Auto-AVSR replace the standard transformer layers with conformer layers, where convolutions are additionally interleaved with self-attention layers. We rely on the publicly available source code and their pre-trained weights for evaluation. 
Overall, the average WER ranges between $19$ and $40$ on the LRS3 test set. Here, \reftab{sota_high_res} shows the main results on both LRS3 and our proposed test sets. Next, we describe the main trends that arise from the experiments.

\noindent\textbf{Significant drops in WER:} All models see a clear drop in WER from the LRS3 to the WildVSR test sets. For the best model on LRS3 (\ie, Auto-AVSR: $19.1$ WER), it loses nearly $20$ points to achieve $38.6$ WER on WildVSR. Also, the self-supervised models pre-trained on 1759h of VoxCeleb2-en + LRS3 are slightly more robust to pre-training on the 433h of LRS3 only (average $21$ \vs $32$ WER drops). In fact, the WER on the WildVSR test set can be linearly approximated given the WER on the LRS3 test set. Let's denote a pair of scores for all models $(wer^{\text{LRS3}}, wer^{\text{WildVSR}})$, then calculate the respective means as: $\mu_{\text{LRS3}} = \frac{1}{M} \sum wer^{\text{LRS3}}_{i}$ and $\mu_{\text{WildVSR}} = \frac{1}{M} \sum wer^{\text{WildVSR}}_{i}$, where $M$ denotes the number of all the models considered. Each score is normalized as: 
\begin{equation}
b = \mu_{\text{WildVSR}} - (m \cdot \mu_{\text{LRS3}}), 
\end{equation}
where $m=\nicefrac{\sigma}{\gamma}$. Here, $\sigma$ and $\gamma$ are computed as
\begin{equation}
\sigma = \sum_{i=1}^{M}(\Delta wer_{i}^{\text{LRS3}} \cdot \Delta wer_{i}^{\text{WildVSR}}),
\end{equation}
\begin{equation}
\gamma = \sum_{i=1}^{M}(\Delta wer_{i}^{\text{LRS3}})^{2},
\end{equation}
where $\Delta wer_i^{k} = | wer_i^{k} - \mu^{k} |$ and $k \in \{\text{LRS3}, \text{WildVSR}\}$.
% 
% \begin{equation}
% \Delta wer_i^{\text{LRS3}} = | wer_i^{\text{LRS3}} - \mu^{\text{LRS3}} |
% \end{equation}
% 
% \begin{equation}
% \Delta wer_i^{\text{Ours}} = | wer_i^{\text{Ours}} - \mu^{\text{Ours}}|
% \end{equation}
% 
% 
% 
% \begin{equation}
% m = \frac{\sigma}{\gamma}
% \end{equation}
% 
% 
Using the WER pairs from \reftab{sota_high_res}, the new WER of a model is approximately given by the following formula:
\begin{equation}
wer^{\text{WildVSR}} \approx 1.31 \cdot wer^{\text{LRS3}}  +14.05 .
\label{linear_wer}
\end{equation}
Since $m {>} 1$, it indicates that models with lower WER on LRS3 experience a comparatively smaller decline when tested on WildVSR. This suggests that better performance on LRS3 yields improved models robustness. We hypothesize that not much overfitting is happening in LRS3, as we do not observe any WER diminishing returns. Hence, there is minimal change in models' ordering across both test sets.

\begin{figure}[t]
    \centering
    \includegraphics[trim={0 0.3cm 0 1.2cm},clip,width=1.00\linewidth]{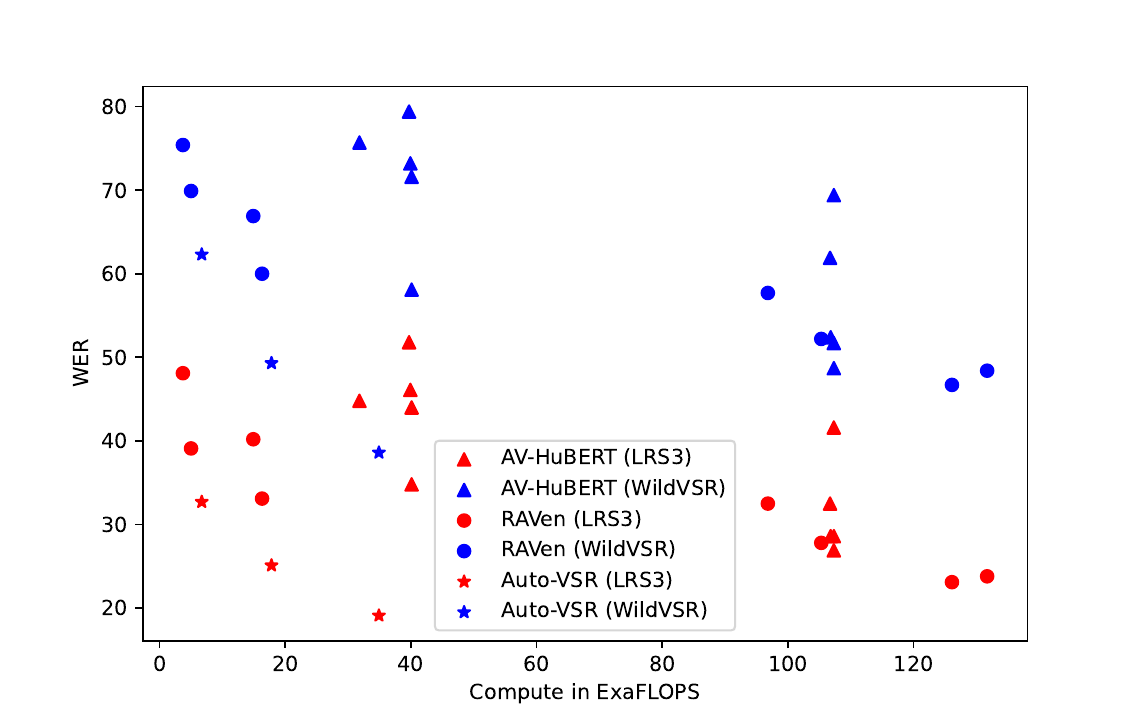}
    \caption{\textbf{Training compute (in exaFLOPs) \vs performance (in WER)}. The best performing models of AV-HuBERT and RAVen that employ pretraining + finetuning achieve only a moderate performance while requiring ${\approx} 3.6\times$ training compute, compared to Auto-AVSR that is trained in a fully-supervised manner.\vspace{-0.2cm}}
    \label{fig:compute}
\end{figure}

\noindent\textbf{Training compute \vs performance:} To estimate the training compute in FLOPs for each model, we utilize the methodologies proposed by \cite{kaplan2020scaling}. This estimation considers various factors including: model size, batch size, training data, and the specifics of the training procedure. As illustrated in \reffig{fig:compute}, supervised models, notably Auto-AVSR, demonstrate an impressive balance between performance and computational efficiency. On the other hand, self-supervised methods (AV-HuBERT and RAVen) demand a significantly higher compute budget ($\approx 3.6\times$ the compute of Auto-AVSR), while achieving only a moderate performance in terms of WER. Further details on the compute calculations are provided in Section~C of supplementary.

\noindent\textbf{Mode collapse:} 
% Figure A.1 illustrates examples from our test set. 
Whilst Wav2vec2.0 fails in matching the exact target speech, the predictions are closer and represent reasonable errors, like homophones (\eg, PARATON  \vs PERITON, LIKE POOR \vs LUDPORE). This suggests that Wav2vec2.0 robustly detect the input patterns, but might miss on prediction due to entangled representations. Clearly however, the failure of the state-of-the-art VSR model, Auto-AVSR, reveals a different nature of error compared to Wav2vec2.0. In the case of Auto-AVSR failure, the predictions deviate significantly from the target speech and often appear to be random. For instance, it produces degenerated predictions, \eg, SPORTING BUSINESS \vs THERE IS BOARDING BUSES. 
% in $1^{st}$ row. 

\subsection{Model Consistency} 
The word error rate (WER) \cite{klakow2002testing} is a standard metric used for comparing different VSR models. However, given that the test set videos have varying target lengths, weighted average WER ($\mu_{wer}$) across the test set might not be sufficient for comparing different approaches, \eg, a model might fit precisely to some samples while having poor predictions for others. Furthermore, we observe that the WER distribution on the LRS3 test set is non-symmetric with more mass around $0$-$20$, while the weighted standard deviation ($\sigma_{wer}$) is in the order of the mean. Thus, we combine both mean and standard deviation in a unified rank metric, as $\mu_{wer}(1+\sigma_{wer})$, to compare the models. Such a metric correctly penalizes models that achieve lower $\mu_{wer}$ at the cost of higher $\sigma_{wer}$. Let us denote $y_{N}$ and $\bar y_{N}$ as the set of ground-truth labels and predictions respectively, where $N$ is the number of samples in the test set. The WER is calculated between each pair $wer_{i} = (y_{i}, \bar y_{i})$, then the set of all scores is denoted as $wer_{N}$. To account for the variable length targets, the average WER is given by:
% 
% \begin{equation}\label{eq1}
%     \begin{split}
%     \mu_{wer} &= \frac{\sum_{i=1}^{N} w_{i} \alpha_{i}}{\sum_{i=1}^{N}  \alpha_{i}} \\
%         &= \sum_{i=1}^{N} \frac{\alpha_{i}}{\sum_{i=1}^{N} \alpha_{i}} w_{i} = \sum_{i=1}^{N} p_{i} w_{i},
%     \end{split}
% \end{equation}
\begin{equation}\label{eq1}
    % \begin{split}
    \mu_{wer} = \frac{\sum_{i=1}^{N} w_{i} \alpha_{i}}{\sum_{i=1}^{N}  \alpha_{i}} \quad = \quad \sum_{i=1}^{N} p_{i} w_{i},
        % &= \sum_{i=1}^{N} \frac{\alpha_{i}}{\sum_{i=1}^{N} \alpha_{i}} w_{i} = \sum_{i=1}^{N} p_{i} w_{i},
    % \end{split}
\end{equation}
where $p_{i} = \frac{\alpha_{i}}{\sum_{i=1}^{N} \alpha_{i}}$ with $\alpha_{i}$ denoting the number of words in $y_{i}$. We define the variance as follows:
\begin{equation}\label{eq2}
    \begin{aligned}
    \sigma_{wer} = \sum_{i=1}^{N} (w_{i} -\mu_{wer})^{2}p_{i}.
    \end{aligned}
\end{equation}
We observe that the WER distribution on the LRS3 test set is non-symmetric, with more mass around $0$, and the standard deviation is in the order of the mean, thus, we propose the following metric to rank the models, based on their standard deviation and mean WER:
\begin{equation}\label{eq3}
    \begin{aligned}
    Rank_{wer} = \mu_{wer}(1+\sigma_{wer}).
    \end{aligned}
\end{equation}
The goal is to have an increasing function in both $\mu_{wer}$ and $\sigma_{wer}$, to penalize models that might have lower mean WER but higher standard deviation.  

\noindent\textbf{Ranking models:} As shown in \reftab{sota_high_res}, models have lower weighted standard deviation on our WildVSR test set ($\sigma_{wer} \approx 0.10$), hence, closer $Rank_{wer}$ compared to $wer$. Surprisingly however, on LRS3, the $\sigma_{wer}$ is in the order of the mean WER for the models ($\sigma_{wer} \approx 0.30$). This suggests that there is a significant amount of variations and inconsistencies in the performance of the models, where the predictions match exactly certain samples from the LRS3 test set, which is not the case for the WildVSR test set. This highlights that VSR models can be sensitive to various factors, such as diverse lip sequences with varying acoustic conditions, accents, vocabulary, or speaking styles. The high $\sigma_{wer}$ suggests that approaches in  \reftab{sota_high_res} are not robust enough to handle such variability, and their performance fluctuate significantly across samples.

\subsection{Analysis of Representation Variability}

\begin{figure}
    \centering
    \includegraphics[trim={0 0.5cm 0 0.1cm},clip,width=1.00\linewidth]{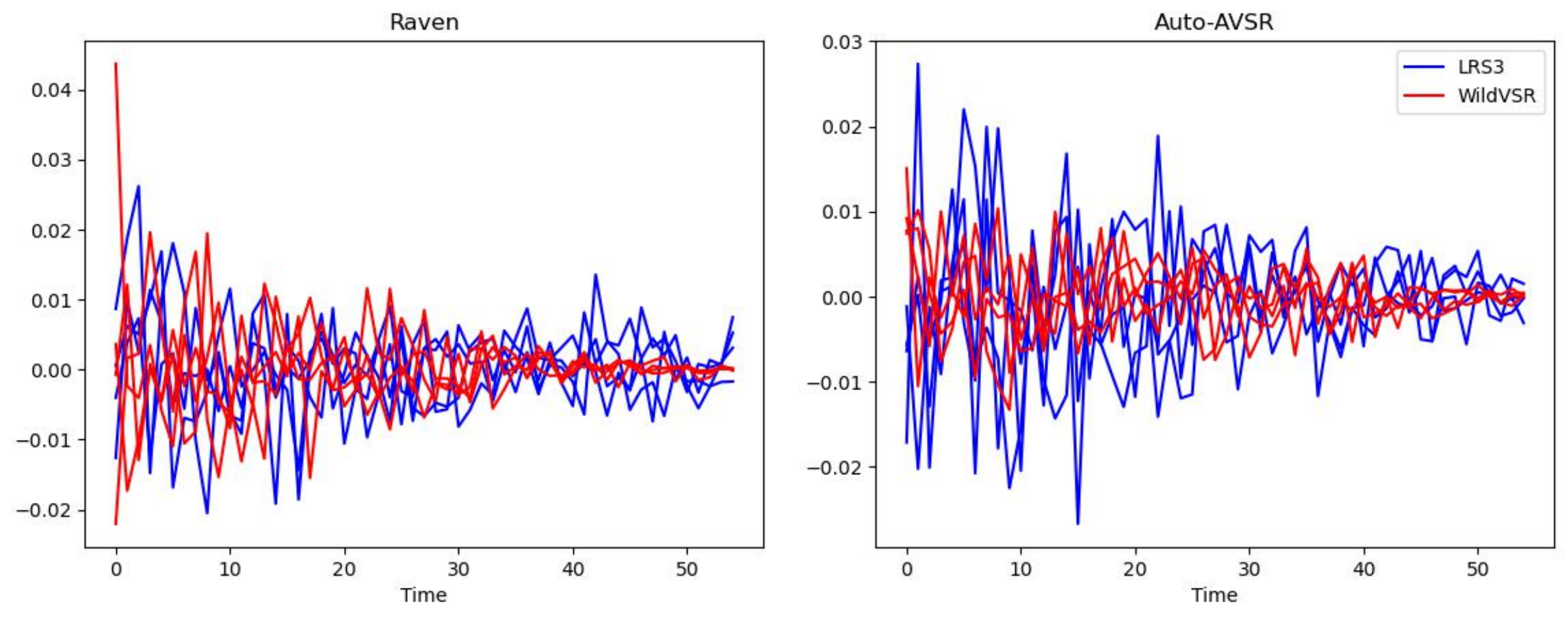}
    \caption{\textbf{Visualization of the dominant spatial modes of the Tucker decomposition over time} of the encoder representations on four sample videos each from LRS3 (in blue) and WildVSR (in red) test sets. The representations are obtained from the best variants of RAVen (on the left) and Auto-AVSR (on the right) models. The detected patterns on LRS3 are salient, whereas fewer modes are detected on WildVSR.\vspace{-0.2cm}}
    \label{fig:tucker}
\end{figure}

With a VSR model denoted as $f_{\theta}$, we sample a batch of video sequences $X_{\text{LRS3}}$ and $X_{\text{WildVSR}}$ from LRS3 and our test sets, respectively. We employ $f_{\theta}$ to map each batch to the representations space $z_{\text{LRS3}}$ and $z_{\text{WildVSR}}$ both $\in \mathbb{R}^{B \times T \times D}$, where $B$ is batch size, $T$ is temporal dimension and $D$ denotes spatial features dimension. To investigate the variability of representations across the two test sets, we employ a combination of power iteration and Tucker decomposition. First, we calculate the power iteration for each representation tensor. The power iteration \cite{mises1929praktische} process allows us to identify the dominant eigenvector in each tensor, which represents the principal direction of variability. The power iteration is performed iteratively until convergence, and then we normalize each representation tensor to be in the same scale. Next, we employ the Tucker decomposition \cite{tucker1966some} to extract the underlying factors of each normalized representation tensor. The Tucker decomposition is a tensor factorization technique that decomposes a tensor into a set of core tensors and factor matrices. For $z_{\text{lrs3}}$ and $z_{\text{WildVSR}}$, the Tucker decomposition is performed with a rank of $(r, r, r)$, resulting in
\begin{equation}
z_{k} = C_{k} \times_1 F_{k}^{(1)} \times_2 F_{k}^{(2)} \times_3 F_{k}^{(3)}, 
\end{equation}
% 
% \begin{equation}
% z_{\text{lrs3}} = C_{\text{lrs3}} \times_1 F_{\text{lrs3}}^{(1)} \times_2 F_{\text{lrs3}}^{(2)} \times_3 F_{\text{lrs3}}^{(3)},
% \end{equation}
% 
% 
% \begin{equation}
% z_{\text{WildVSR}} = C_{\text{lrs3}} \times_1 F_{\text{lrs3}}^{(1)} \times_2 F_{\text{lrs3}}^{(2)} \times_3 F_{\text{lrs3}}^{(3)}.
% \end{equation}
% 
% 
where $k \!\in\! \{\text{LRS3}, \text{WildVSR}\}$. The factor matrices obtained from Tucker decomposition represent the latent factors of variability in the representation tensors. Let $F_{\text{LRS3}}^{(i)}$ and $F_{\text{WildVSR}}^{(i)}$ denote the factor matrices corresponding to $z_{\text{LRS3}}$ and $z_{\text{WildVSR}}$, respectively.
To project each representation tensor onto the factor matrices, we perform a multi-mode dot product. This operation aligns the representation tensor with the corresponding factor matrices, capturing the influence of each factor on the tensor. With $k \in \{\text{LRS3}, \text{WildVSR}\}$, the projection is done as follows:
\begin{equation}
\text{proj}_{k} = z_{k} \times_1 F_{k}^{(1)^T} \times_2 F_{k}^{(2)^T} \times_3 F_{k}^{(3)^T}.
\end{equation}
% 
% \begin{equation}
% \text{proj}_{\text{WildVSR}} = z_{\text{WildVSR}} \times_1 F_{\text{WildVSR}}^{(1)^T} \times_2 F_{\text{WildVSR}}^{(2)^T} \times_3 F_{\text{WildVSR}}^{(3)^T},
% \end{equation}
% % 
% % 
% \begin{equation}
% \text{proj}_{\text{lrs3}} = z_{\text{lrs3}} \times_1 F_{\text{lrs3}}^{(1)^T} \times_2 F_{\text{WildVSR}}^{(2)^T} \times_3 F_{\text{lrs3}}^{(3)^T}.
% \end{equation}
% 
Finally, we visualize the projected tensors $\text{proj}_{\text{WildVSR}}$ and $\text{proj}_{\text{lrs3}}$ to examine the variability of representations across the two test sets. This visualization provides insights into the differences in the learned representations and helps assess the effectiveness of our model. The entire procedure enables a comprehensive analysis of the variability in representations across different test sets, facilitating the evaluation and interpretation of model performance.

\begin{figure}
    \centering
    \includegraphics[trim={0 0.5cm 0 0.1cm},clip, width=1.00\linewidth]{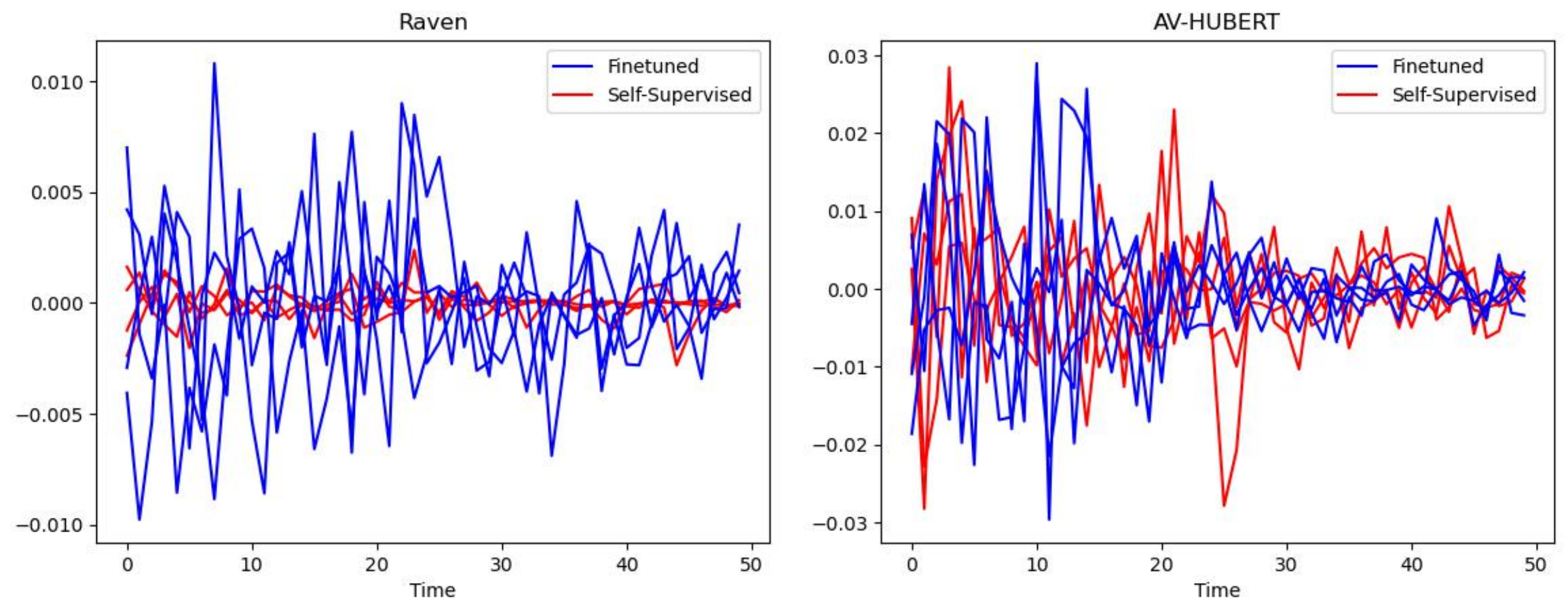}
    \caption{\textbf{Visualization of Tucker decomposition's spatial mode over time} for RAVen (on the left) and AV-HuBERT (on the right) encoder representations on sample videos from LRS3, before (in red) and after (in blue) finetuning. We see that RAVen encoder before finetuning fails to capture salient modes indicating that VSR representations are better learned only during its finetuning stage.\vspace{-0.2cm}}
    \label{fig:tucker_ssl}
\end{figure}

\noindent\textbf{Evaluation:} We select a batch of 128 examples from both LRS3 and our WildVSR test sets, and use the best variants of RAVen and Auto-AVSR for the analysis. \reffig{fig:tucker} shows the projected tensors over time on both test sets. Observing the visualizations, it becomes evident that both models exhibit a considerably stronger response to the sequences presented in the LRS3 test set. These models successfully capture the underlying modes and patterns inherent in the input signals, enabling accurate detection and subsequent generation of precise transcriptions. The rich and diverse set of features recognized by the models within the LRS3 test set contributes to their superior predictive capabilities. Conversely, when examining the representations derived from our own test set, a notable difference emerges. The features extracted from our test set display a distinct lack of variability, indicating a reduced number of patterns being recognized by the models. Consequently, the predictive performance of the models on our test set is compromised, leading to sub-optimal transcription outcomes.

\noindent\textbf{Which SSL is better?} We notice fundamental differences between RAVen and AV-HuBERT self-supervised procedures, where the former learns two separate encoders for video and audio modalities and optimizes to match the latents. Differently, the latter fuses them in a single encoder and learns to predict pre-assigned cluster memberships. Hence, we use the same Tucker decomposition to compare their representations before and after finetuning. As shown in \reffig{fig:tucker_ssl}, SSL RAVen encodes much less information explaining the need for requiring 75 epochs at for finetuning stage. In contrast,  AV-HuBERT detects modes even without seeing any labelled data, suggesting the reason behind the small finetuning budget. We hypothesize that coarse-grained clustering as done in AV-HuBERT better suits the VSR task, as it pushes the encoder to learn phonemes that can be lightly mapped to word labels.

\begin{table}[t]

\caption[Caption for LOF]{\textbf{Performance comparison on multiple folds of our test set.} The folds are the intersection of where all models are below $30$ WER for \textit{top}-k, and more than $50$ WER for \textit{bottom}-k. We observe that the models match their LRS3 performance on \textit{top}-k examples.\vspace{-0.2cm}}
\centering
\setlength{\tabcolsep}{14pt}
     \adjustbox{width=1\columnwidth}{

\begin{tabular}{lrrr}

\toprule
   \textbf{Model} &\multicolumn{3}{c}{Video folds} \\
    \cline{2-4}
     & Top-k & Bottom-k & All  \\
\midrule
 Auto-AVSR~\cite{ma2023auto} & 18.6 & 71.4 & 38.6  \\
 AV-HuBERT~\cite{shi2022learning} w/ self-training & 24.8 & 77.3 & 48.7  \\
 RAVen~\cite{haliassos2022jointly} w/ self-training & 23.5 & 75.0 & 46.7 \\

\bottomrule
\vspace{-0.5cm}
\end{tabular}
}
\label{folds_results}
\end{table}

\begin{figure}[t]
    \centering
    \includegraphics[trim={0 0.4cm 0 0.1cm},clip,width=1.00\linewidth]{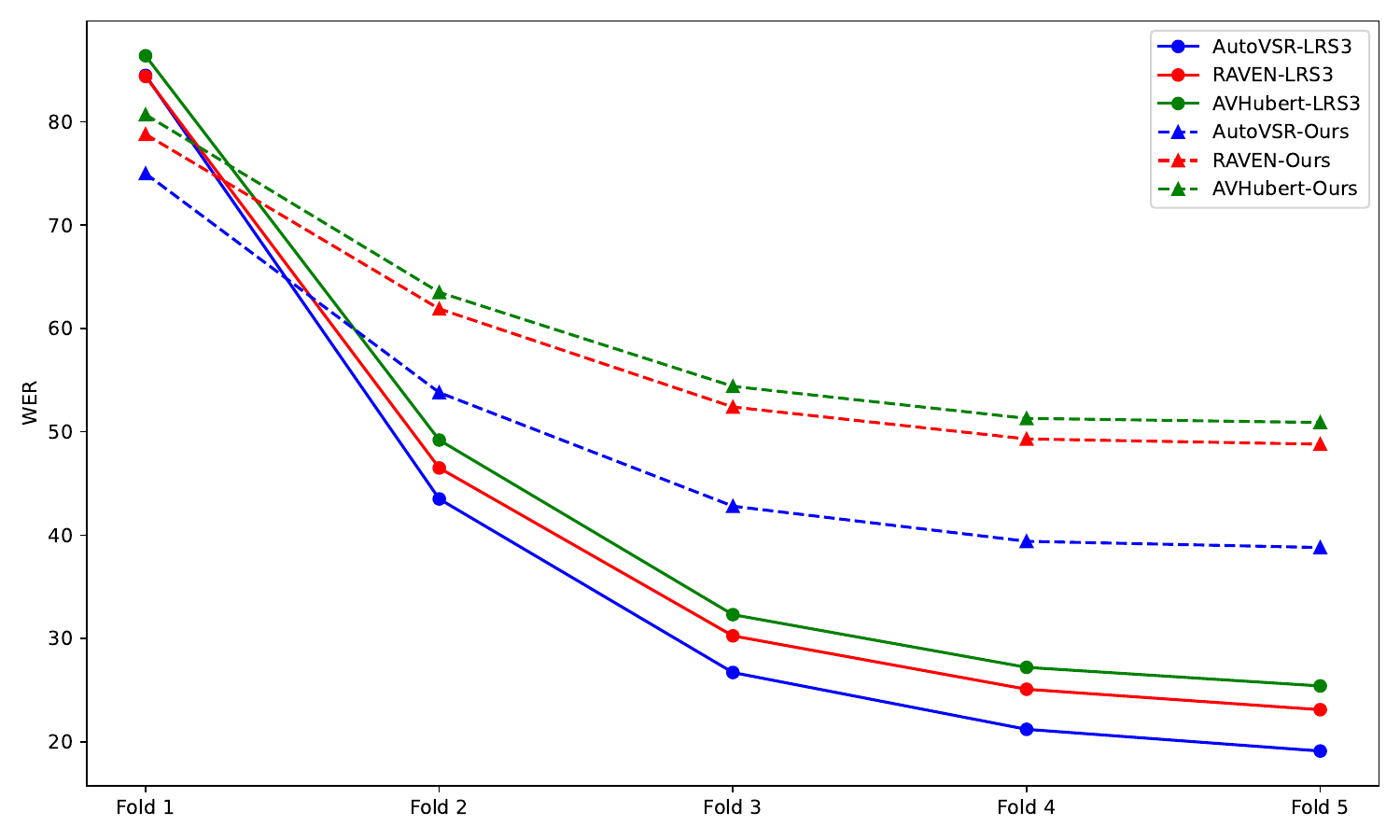}\vspace{-0.1cm}
    \caption{\textbf{Performance (WER) comparison across five different data folds}. Fold $1$ is an overlap subset where all models obtain a WER score higher than $50$. We keep progressively adding data from the remaining test set to create Folds $2$ to $5$, where Fold $5$ contains all the test set videos.\vspace{-0.3cm}}
    \label{fig:wer_folds}
\end{figure}

\section{Potential Causes for WER drop}

Following \cite{recht2019imagenet}, the error difference between the respective test sets can be decomposed into three parts:
\begin{equation}
\resizebox{0.9\linewidth}{!}{$\mathcal{L}_{WildVSR} - \mathcal{L}_{lrs3} = \underbrace{(\mathcal{L}_{WildVSR} - \mathcal{L}_{lrs3})}_{\text{Adaptivity gap}} + \underbrace{(\mathcal{L}_{WildVSR} - \mathcal{L}_{lrs3})}_{\text{Distribution Gap}} + \underbrace{(\mathcal{L}_{WildVSR} - \mathcal{L}_{lrs3})}_{\text{Generalization gap}}$}
\end{equation}
The \textit{Adaptivity gap} quantifies the extent to which adjusting (adapting) a model to fit the specific test set $\mathcal{L}_{lrs3}$ leads to an underestimation of the test error. The \textit{Distribution gap} measures how our new data distribution and generation process affects the model performance. The \textit{Generalization gap} is influenced by the inherent random sampling error. These components are hard to track, and distinguish in practice.

\noindent\textbf{Linear proportional gains:} \refeq{linear_wer} suggests that no diminishing returns is apparent. This strongly indicates that the reduction in Word Error Rate (WER) could primarily be attributed to the \textit{Distribution Gap}. Despite our best efforts to emulate the original LRS3 dataset creation procedure, the distribution gap remains the main explanation for the observed decreases in WER. Upon examination of our data \reftab{folds_results}, we observe that a subset of our test set exhibits similar complexity levels to those found in LRS3, thus models achieve their reported LRS3 WER scores. We postulate that our test set includes a larger number of examples from difficult LRS3 modes, in addition to new modes not found in LRS3. While current VSR models showcase impressive WER scores on the original LRS3 test set,  they still encounter difficulties in effectively generalizing from `easy' lip sequences to more challenging ones.

\begin{table}[t] 
    \caption{\textbf{Performance comparison on our proposed test set with varying the attributes}. We report the best variant per model.\vspace{-0.3cm}}
    \centering
    \setlength{\tabcolsep}{10pt}
    
    \adjustbox{width=\columnwidth}{
    \begin{tabular}
    { l c c | c c} 
     \toprule[0.1em]
      \textbf{Method} &  \multicolumn{2}{c|}{\textbf{Accent}} & \multicolumn{2}{c}{\textbf{Gender}} \\
    % \cline{2-5}
     & Native & Non-Native  & Male & Female \\

    \midrule
    
    Auto-AVSR~\cite{ma2023auto} & 35.1 & 46.9 & 38.3 & 37.4 \\

    AV-HuBERT~\cite{shi2022learning} & 47.5 & 55.3 & 49.2 & 47.7   \\
   RAVen~\cite{haliassos2022jointly}  & 45.2 & 55.0 & 47.5 & 45.3 \\
    
     \bottomrule[0.1em]
    \end{tabular}
    }
    \label{attributes}
    \vspace{-0.2cm}
    
\end{table}

\begin{figure}[t]
    \centering
    \includegraphics[trim={0 0.4cm 0 0.1cm},clip,width=1.00\linewidth]{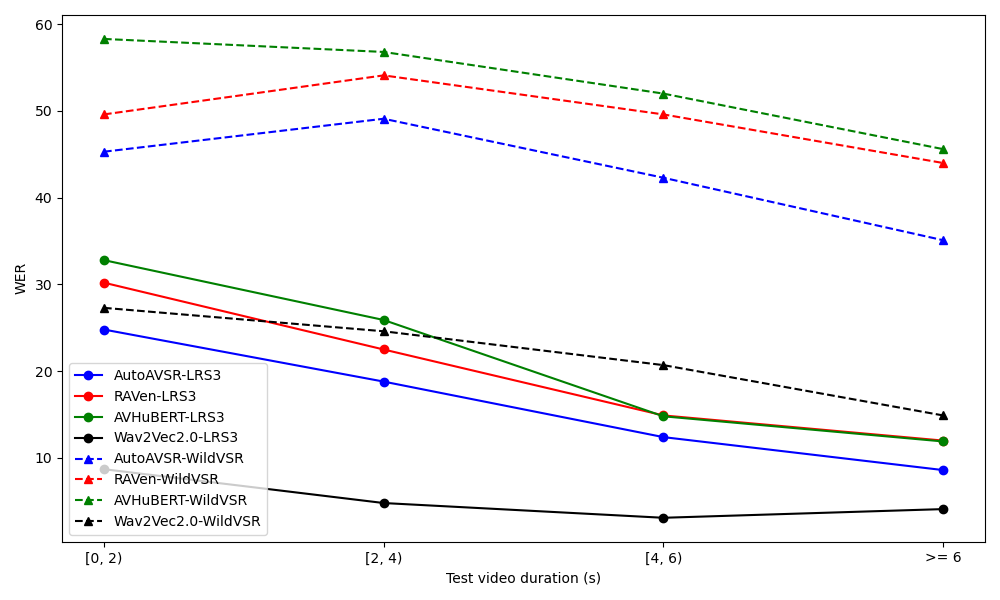}\vspace{-0.1cm}
    \caption{\textbf{Variation of WER as duration of video clips is varied}. VSR models decode better with longer context (video duration).\vspace{-0.2cm}}
    \label{fig:duration}
\end{figure}

\noindent\textbf{Hard samples:} As shown in \reffig{fig:wer_folds}, we selected a data subset where all models obtain a WER score higher than 50. The subsets contain 110/639 samples for LRS3/WildVSR respectively. On both LRS3 and WildVSR, models obtain a WER $\ge$ 75 for these challenging examples. We checked these samples manually, and noticed a high variability in head poses along with shorter videos. There might be other confounding variables from the accents/vocabulary that are harder to assess. \reffig{fig:duration} shows the performance comparison on different folds obtained by partitioning the test sets based on the lip sequence length. We observe that shorter videos (less than $2$ seconds, \ie, $50$ frames) present a bottleneck, which results in performance degradation of the approaches from their corresponding average WER on the whole test sets. This is likely due to the lack of rich contextual features in shorter video sequences, which leads to sub-optimal temporal modeling in the video encoder. Furthermore, we kept progressively adding samples from the remaining test set, and observed a faster decay on LRS3. We hypothesise that the higher amount  of `hard samples' (23\%) in WildVSR (\vs 08\% for LRS3) is responsible for this behavior as the final WER is averaged across samples.

\noindent\textbf{Speaker characteristics:} As demonstrated in Table \ref{attributes}, the models show an average decrease of 8 points when evaluated on non-native speakers compared to native speakers. Additionally, there is a relatively similar performance between male and female speakers across the models. Also see Table.~A.3 for the scores on more attributes.

\noindent\textbf{Head poses:} Here, we create two subsets $\mathcal{S}_{1}$ and $\mathcal{S}_{2}$ with $1010$ and $639$ videos, respectively. $\mathcal{S}_{1}$ contains the best-performing videos across different models, while $\mathcal{S}_{2}$ contains videos that are hard to decode for all the models. Next, we detect sequences with frontal and extreme poses by first recovering the 3D head pose using \cite{filntisis2022visual} and then regressing the 3D-model parameters that best fit to each image frame using a parametric 3D model \cite{Li2017LearningScans} learned from 3D scans of human faces. Frontal and extreme poses are considered based on predefined face angles. We perform this analysis and find that extreme poses represent $31\%$ and  $52\%$ on  $\mathcal{S}_{1}$ and $\mathcal{S}_{2}$, respectively. This suggests that extreme head poses adversely affect the model performance for VSR task.

\noindent\textbf{Vocabulary:} As demonstrated in \reftab{folds_results}, state-of-the-art (SoTA) models reproduce their reported scores on a subset of our test set (specifically, \textit{top}-k, comprising $1010$ videos). We conducted an in-depth analysis of the vocabulary utilized in this fold and found that it comprises a total of $2315$ words. Interestingly, this word set exhibits a $75\%$ similarity to the LRS3 vocabulary, which consists of $1969$ words. This observation may suggest that VSR models tend to perform optimally when operating within a constrained vocabulary set, analogous to the one present in the LRS3 dataset. This underlines the significant influence that vocabulary consistency and word choice have on the performance of VSR models. The restricted vocabulary environment may simplify the model's task, leading to a higher performance. Understanding these limitations can provide crucial insights for future development of VSR models.

\noindent\textbf{Unified model architectures:} Despite the many approaches in VSR literature, they mostly share the same architecture  while differing in the training procedures and objectives. The inductive bias-free nature of transformers helps preventing the explicit definition of \textit{overfitting}. It should be noted that while these methods do eliminate the most salient aspects of overfitting, they do not eliminate all possibilities. There exists a scenario where any degree of test set adaptivity results in a consistent decrease in accuracy across all models. In such a case, the diminishing returns phenomenon would not be observed since subsequent models could still exhibit improved performance. Detecting and studying this specific manifestation of adaptive overfitting would likely necessitate the use of a novel test set that truly adheres to independent and identically distributed properties, rather than being derived from a distinct data collection effort. Identifying an appropriate dataset for conducting such an experiment remains extremely challenging.

\section{Discussion}

\noindent\textbf{Recommendations:} As observed earlier in compute-performance trade-off section, although self-supervised learning followed by fine-tuning paradigm might seem exciting, it emerges as a less optimal approach for VSR. The inherent complexity of video data intensifies the computational demands, making it an expensive avenue to explore. Moreover, as these learned encoders are utilized for a singular downstream task of VSR, the potential benefits of SSL are somewhat neutralized. Additionally, the fintetuning phase can consume a budget comparable to that of direct supervised training, as in RAVen~\cite{haliassos2022jointly}. \reffig{fig:compute} shows the effectiveness of fully-supervised learning when supplemented with state-of-the-art ASR models, such as Whisper \cite{radford2022robust}, serving as automatic labelers. This methodology, as proved by Auto-AVSR~\cite{ma2023auto}, provides a good balance between performance and computational efficiency, while still achieving superior performance.
% This strategy not only conserves computational resources but also ensures superior model performance.

\noindent\textbf{Ethical considerations:} Due to the data collection process focusing on YouTube, biases inherent to the platform may be present in the test set. Also, while measures are taken to ensure diversity in content, the dataset might still be skewed towards certain types of content due to the filtering process. Furthermore, we have taken specific steps to ensure that WildVSR respects individual privacy. Firstly, since most VSR approaches utilize only the $96 \times 96$ cropped region around the mouth as input, we make available the cropped sequence to reduce the potential for individual identification, emphasizing only the pertinent region for VSR model's input. Also, we  provide a mechanism for individuals who recognize themselves in the test set to opt-out. Should someone wish to have their data removed, they can contact us and we will promptly exclude their content.

\noindent\textbf{Conclusion:} In this work, we have highlighted the lack of generalization within the field of Visual Speech Recognition (VSR) due to an excessive focus on the Lip Reading Sentences-3 (LRS3) test set. To mitigate this, we have proposed a new VSR test set, named WildVSR, incorporating a higher visual diversity and spoken vocabulary. Indeed, the benchmarking of a wide range of publicly available VSR models on this new test set revealed significant drops in performance compared to the LRS3 test set. This outcome underlines the models' difficulties in generalizing to “harder” lip sequences present in our test set. Interestingly, the comparative ranking of models remained consistent across the original and new test sets, indicating that these performance drops are not merely a result of over-tuning for specific lip sequences on LRS3 test set. We also introduced a novel metric that combines the mean and standard deviation of Word Error Rates (WER), better capturing a model's consistency across various test samples. It is our hope that this will stimulate the development of more robust VSR models, furthering advancements in this challenging field.

% Furthermore, we aim to publicly release our new test benchmark, providing an enriched dataset for future VSR research. It is our hope that this will stimulate the development of more robust VSR models, furthering advancements in this challenging field.

{\small
\bibliographystyle{ieee_fullname}
\bibliography{egbib}

\begin{thebibliography}{10}\itemsep=-1pt

\bibitem{afouras2018lrs3}
Triantafyllos Afouras, Joon~Son Chung, and Andrew Zisserman.
\newblock Lrs3-ted: a large-scale dataset for visual speech recognition.
\newblock {\em arXiv preprint arXiv:1809.00496}, 2018.

\bibitem{Afouras2019ASRReading}
Triantafyllos Afouras, Joon~Son Chung, and Andrew Zisserman.
\newblock {ASR is all you need: cross-modal distillation for lip reading}.
\newblock {\em ICASSP, IEEE International Conference on Acoustics, Speech and Signal Processing - Proceedings}, 2020-May:2143--2147, 11 2019.

\bibitem{baevski2020wav2vec}
Alexei Baevski, Yuhao Zhou, Abdelrahman Mohamed, and Michael Auli.
\newblock wav2vec 2.0: A framework for self-supervised learning of speech representations.
\newblock {\em Advances in neural information processing systems}, 33:12449--12460, 2020.

\bibitem{chang2023conformers}
Oscar Chang, Hank Liao, Dmitriy Serdyuk, Ankit Shah, and Olivier Siohan.
\newblock Conformers are all you need for visual speech recogntion.
\newblock {\em arXiv preprint arXiv:2302.10915}, 2023.

\bibitem{chollet2019measure}
Fran{\c{c}}ois Chollet.
\newblock On the measure of intelligence.
\newblock {\em arXiv preprint arXiv:1911.01547}, 2019.

\bibitem{chung2018voxceleb2}
Joon~Son Chung, Arsha Nagrani, and Andrew Zisserman.
\newblock Voxceleb2: Deep speaker recognition.
\newblock {\em arXiv preprint arXiv:1806.05622}, 2018.

\bibitem{chung2017lip}
Joon~Son Chung and Andrew Zisserman.
\newblock Lip reading in the wild.
\newblock In {\em Computer Vision--ACCV 2016: 13th Asian Conference on Computer Vision, Taipei, Taiwan, November 20-24, 2016, Revised Selected Papers, Part II 13}, pages 87--103. Springer, 2017.

\bibitem{chung2017out}
Joon~Son Chung and Andrew Zisserman.
\newblock Out of time: automated lip sync in the wild.
\newblock In {\em Computer Vision--ACCV 2016 Workshops: ACCV 2016 International Workshops, Taipei, Taiwan, November 20-24, 2016, Revised Selected Papers, Part II 13}, pages 251--263. Springer, 2017.

\bibitem{ephrat2018looking}
Ariel Ephrat, Inbar Mosseri, Oran Lang, Tali Dekel, Kevin Wilson, Avinatan Hassidim, William~T Freeman, and Michael Rubinstein.
\newblock Looking to listen at the cocktail party: A speaker-independent audio-visual model for speech separation.
\newblock {\em arXiv preprint arXiv:1804.03619}, 2018.

\bibitem{fefferman2016testing}
Charles Fefferman, Sanjoy Mitter, and Hariharan Narayanan.
\newblock Testing the manifold hypothesis.
\newblock {\em Journal of the American Mathematical Society}, 29(4):983--1049, 2016.

\bibitem{filntisis2022visual}
Panagiotis~P Filntisis, George Retsinas, Foivos Paraperas-Papantoniou, Athanasios Katsamanis, Anastasios Roussos, and Petros Maragos.
\newblock Visual speech-aware perceptual 3d facial expression reconstruction from videos.
\newblock {\em arXiv preprint arXiv:2207.11094}, 2022.

\bibitem{7040826}
Igor~S. Gruzman and Anna~S. Kostenkova.
\newblock Algorithm of scene change detection in a video sequence based on the threedimensional histogram of color images.
\newblock In {\em 2014 12th International Conference on Actual Problems of Electronics Instrument Engineering (APEIE)}, pages 1--1, 2014.

\bibitem{haliassos2022jointly}
Alexandros Haliassos, Pingchuan Ma, Rodrigo Mira, Stavros Petridis, and Maja Pantic.
\newblock Jointly learning visual and auditory speech representations from raw data.
\newblock {\em arXiv preprint arXiv:2212.06246}, 2022.

\bibitem{jaiswal2020survey}
Ashish Jaiswal, Ashwin~Ramesh Babu, Mohammad~Zaki Zadeh, Debapriya Banerjee, and Fillia Makedon.
\newblock A survey on contrastive self-supervised learning.
\newblock {\em Technologies}, 9(1):2, 2020.

\bibitem{kaplan2020scaling}
Jared Kaplan, Sam McCandlish, Tom Henighan, Tom~B Brown, Benjamin Chess, Rewon Child, Scott Gray, Alec Radford, Jeffrey Wu, and Dario Amodei.
\newblock Scaling laws for neural language models.
\newblock {\em arXiv preprint arXiv:2001.08361}, 2020.

\bibitem{klakow2002testing}
Dietrich Klakow and Jochen Peters.
\newblock Testing the correlation of word error rate and perplexity.
\newblock {\em Speech Communication}, 38(1-2):19--28, 2002.

\bibitem{kosaka2003neural}
Hirotaka Kosaka, Masao Omori, Tetsuya Iidaka, Tetsuhito Murata, T Shimoyama, Tomohisa Okada, Norihiro Sadato, Yoshiharu Yonekura, and Yuji Wada.
\newblock Neural substrates participating in acquisition of facial familiarity: an fmri study.
\newblock {\em Neuroimage}, 20(3):1734--1742, 2003.

\bibitem{Li2017LearningScans}
Tianye Li, Timo Bolkart, Michael~J. Black, Hao Li, and Javier Romero.
\newblock {Learning a model of facial shape and expression from 4D scans}.
\newblock {\em ACM Transactions on Graphics (TOG)}, 36(6), 11 2017.

\bibitem{liu2023synthvsr}
Xubo Liu, Egor Lakomkin, Konstantinos Vougioukas, Pingchuan Ma, Honglie Chen, Ruiming Xie, Morrie Doulaty, Niko Moritz, Jachym Kolar, Stavros Petridis, et~al.
\newblock Synthvsr: Scaling up visual speech recognition with synthetic supervision.
\newblock In {\em Proceedings of the IEEE/CVF Conference on Computer Vision and Pattern Recognition}, pages 18806--18815, 2023.

\bibitem{ma2023auto}
Pingchuan Ma, Alexandros Haliassos, Adriana Fernandez-Lopez, Honglie Chen, Stavros Petridis, and Maja Pantic.
\newblock Auto-avsr: Audio-visual speech recognition with automatic labels.
\newblock In {\em ICASSP 2023-2023 IEEE International Conference on Acoustics, Speech and Signal Processing (ICASSP)}, pages 1--5. IEEE, 2023.

\bibitem{ma2021lira}
Pingchuan Ma, Rodrigo Mira, Stavros Petridis, Bj{\"o}rn~W Schuller, and Maja Pantic.
\newblock Lira: Learning visual speech representations from audio through self-supervision.
\newblock {\em arXiv preprint arXiv:2106.09171}, 2021.

\bibitem{ma2021end}
Pingchuan Ma, Stavros Petridis, and Maja Pantic.
\newblock End-to-end audio-visual speech recognition with conformers.
\newblock In {\em ICASSP 2021-2021 IEEE International Conference on Acoustics, Speech and Signal Processing (ICASSP)}, pages 7613--7617. IEEE, 2021.

\bibitem{ma2022visual}
Pingchuan Ma, Stavros Petridis, and Maja Pantic.
\newblock Visual speech recognition for multiple languages in the wild.
\newblock {\em Nature Machine Intelligence}, pages 1--10, 2022.

\bibitem{ma2022training}
Pingchuan Ma, Yujiang Wang, Stavros Petridis, Jie Shen, and Maja Pantic.
\newblock Training strategies for improved lip-reading.
\newblock In {\em ICASSP 2022-2022 IEEE International Conference on Acoustics, Speech and Signal Processing (ICASSP)}, pages 8472--8476. IEEE, 2022.

\bibitem{Ma2021ContrastiveRepresentations}
Shuang Ma, Zhaoyang Zeng, Daniel McDuff, and Yale Song.
\newblock {Contrastive Learning of Global-Local Video Representations}.
\newblock {\em Advances in Neural Information Processing Systems}, 9:7025--7040, 4 2021.

\bibitem{makino2019recurrent}
Takaki Makino, Hank Liao, Yannis Assael, Brendan Shillingford, Basilio Garcia, Otavio Braga, and Olivier Siohan.
\newblock Recurrent neural network transducer for audio-visual speech recognition.
\newblock In {\em 2019 IEEE automatic speech recognition and understanding workshop (ASRU)}, pages 905--912. IEEE, 2019.

\bibitem{mises1929praktische}
RV Mises and Hilda Pollaczek-Geiringer.
\newblock Praktische verfahren der gleichungsaufl{\"o}sung.
\newblock {\em ZAMM-Journal of Applied Mathematics and Mechanics/Zeitschrift f{\"u}r Angewandte Mathematik und Mechanik}, 9(1):58--77, 1929.

\bibitem{prajwal2022sub}
KR Prajwal, Triantafyllos Afouras, and Andrew Zisserman.
\newblock Sub-word level lip reading with visual attention.
\newblock In {\em Proceedings of the IEEE/CVF Conference on Computer Vision and Pattern Recognition}, pages 5162--5172, 2022.

\bibitem{qi2023yolo5face}
Delong Qi, Weijun Tan, Qi Yao, and Jingfeng Liu.
\newblock Yolo5face: Why reinventing a face detector.
\newblock In {\em Computer Vision--ECCV 2022 Workshops: Tel Aviv, Israel, October 23--27, 2022, Proceedings, Part V}, pages 228--244. Springer, 2023.

\bibitem{radford2022robust}
Alec Radford, Jong~Wook Kim, Tao Xu, Greg Brockman, Christine McLeavey, and Ilya Sutskever.
\newblock Robust speech recognition via large-scale weak supervision.
\newblock {\em arXiv preprint arXiv:2212.04356}, 2022.

\bibitem{recht2019imagenet}
Benjamin Recht, Rebecca Roelofs, Ludwig Schmidt, and Vaishaal Shankar.
\newblock Do imagenet classifiers generalize to imagenet?
\newblock In {\em International conference on machine learning}, pages 5389--5400. PMLR, 2019.

\bibitem{serdyuk2021audio}
Dmitriy Serdyuk, Otavio Braga, and Olivier Siohan.
\newblock Audio-visual speech recognition is worth 32x32x8 voxels.
\newblock In {\em 2021 IEEE Automatic Speech Recognition and Understanding Workshop (ASRU)}, pages 796--802. IEEE, 2021.

\bibitem{serengil2020lightface}
Sefik~Ilkin Serengil and Alper Ozpinar.
\newblock Lightface: A hybrid deep face recognition framework.
\newblock In {\em 2020 Innovations in Intelligent Systems and Applications Conference (ASYU)}, pages 23--27. IEEE, 2020.

\bibitem{serengil2021lightface}
Sefik~Ilkin Serengil and Alper Ozpinar.
\newblock Hyperextended lightface: A facial attribute analysis framework.
\newblock In {\em 2021 International Conference on Engineering and Emerging Technologies (ICEET)}, pages 1--4. IEEE, 2021.

\bibitem{sheng2021cross}
Changchong Sheng, Matti Pietik{\"a}inen, Qi Tian, and Li Liu.
\newblock Cross-modal self-supervised learning for lip reading: When contrastive learning meets adversarial training.
\newblock In {\em Proceedings of the 29th ACM International Conference on Multimedia}, pages 2456--2464, 2021.

\bibitem{shi2022learning}
Bowen Shi, Wei-Ning Hsu, Kushal Lakhotia, and Abdelrahman Mohamed.
\newblock Learning audio-visual speech representation by masked multimodal cluster prediction.
\newblock {\em arXiv preprint arXiv:2201.02184}, 2022.

\bibitem{shi2022robust}
Bowen Shi, Wei-Ning Hsu, and Abdelrahman Mohamed.
\newblock Robust self-supervised audio-visual speech recognition.
\newblock {\em arXiv preprint arXiv:2201.01763}, 2022.

\bibitem{son2017lip}
Joon Son~Chung, Andrew Senior, Oriol Vinyals, and Andrew Zisserman.
\newblock Lip reading sentences in the wild.
\newblock In {\em Proceedings of the IEEE conference on computer vision and pattern recognition}, pages 6447--6456, 2017.

\bibitem{summerfield2022natural}
Christopher Summerfield.
\newblock {\em Natural General Intelligence: How understanding the brain can help us build AI}.
\newblock Oxford University Press, 2022.

\bibitem{todorov2012role}
Alexander Todorov.
\newblock The role of the amygdala in face perception and evaluation.
\newblock {\em Motivation and Emotion}, 36:16--26, 2012.

\bibitem{tucker1966some}
Ledyard~R Tucker.
\newblock Some mathematical notes on three-mode factor analysis.
\newblock {\em Psychometrika}, 31(3):279--311, 1966.

\bibitem{zhu2022vatlm}
Qiushi Zhu, Long Zhou, Ziqiang Zhang, Shujie Liu, Binxing Jiao, Jie Zhang, Lirong Dai, Daxin Jiang, Jinyu Li, and Furu Wei.
\newblock Vatlm: Visual-audio-text pre-training with unified masked prediction for speech representation learning.
\newblock {\em arXiv preprint arXiv:2211.11275}, 2022.

\end{thebibliography}
}

\clearpage

\appendix

\setcounter{table}{0}
\setcounter{figure}{0}
\renewcommand{\thetable}{A.\arabic{table}}
\renewcommand{\thefigure}{A.\arabic{figure}}

\section*{\LARGE Appendices}

In this supplementary, we present additional analysis related to our proposed test set WildVSR. Additional details regarding the data collection are given in Section~\ref{sec:data_collect} followed by additional results on ASR models and word-level lip-reading in Section~\ref{sec:add_results}. Finally, we present additional details regarding the training budget calculation for the VSR models in Section~\ref{sec:training_compute}.

\section{Additional Details on Data Collection\label{sec:data_collect}}

\noindent\textbf{Keywords selection.} In the course of our study,  the tool was configured to systematically extract video IDs from YouTube, utilizing a predefined array of significant keywords as the fundamental criteria for data selection. These keywords, derived from diverse trending and popular thematic categories, directed the tool to gather data across a wide spectrum of content. These keywords are: \textit{knowledge}, \textit{history}, \textit{conference}, \textit{beauty}, \textit{dialogue}, \textit{news}, \textit{talk}, \textit{interview}, \textit{sport}, \textit{health}, \textit{technology}, \textit{conversation}, \textit{cooking}, \textit{lesson}, \textit{tips}, \textit{reading}, \textit{challenges}, \textit{travel}, \textit{course}, \textit{games}. As such, this comprehensive dataset provided a rich substrate for our subsequent investigations, enabling a deeper understanding of the various parameters that govern the content popularity on the platform.

\begin{figure}[t]
    \centering
    \includegraphics[width=1.00\linewidth]{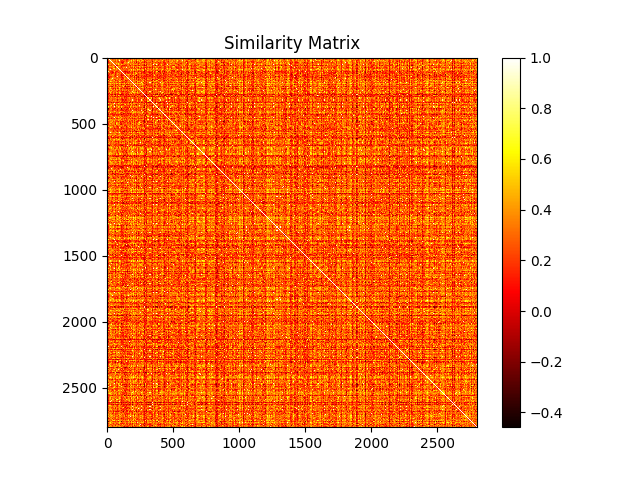}
    \caption{The similarity matrix of the face embeddings using VGG-Face across the test set samples. It can be seen that the similairy is low across our test set, showing the diversity.}
    \label{fig:similarity}
\end{figure}

\noindent\textbf{Similarity measures.} As shown in \reffig{fig:similarity}, the similarity across our test set is relatively low (visually represented by darker colors in the matrix) signifing that the embeddings produced by the VGG-Face model are notably distinct for different test set images. This in turn implies that the VGG-Face model has successfully captured a wide array of facial feature representations, making it capable of distinguishing between different individuals effectively. Moreover, the high diversity within the similarity matrix also indicates that the models are fairly tested without a specific focus on a given facial category. 

\begin{figure*}[t]
    \centering
    \includegraphics[width=1.00\linewidth]{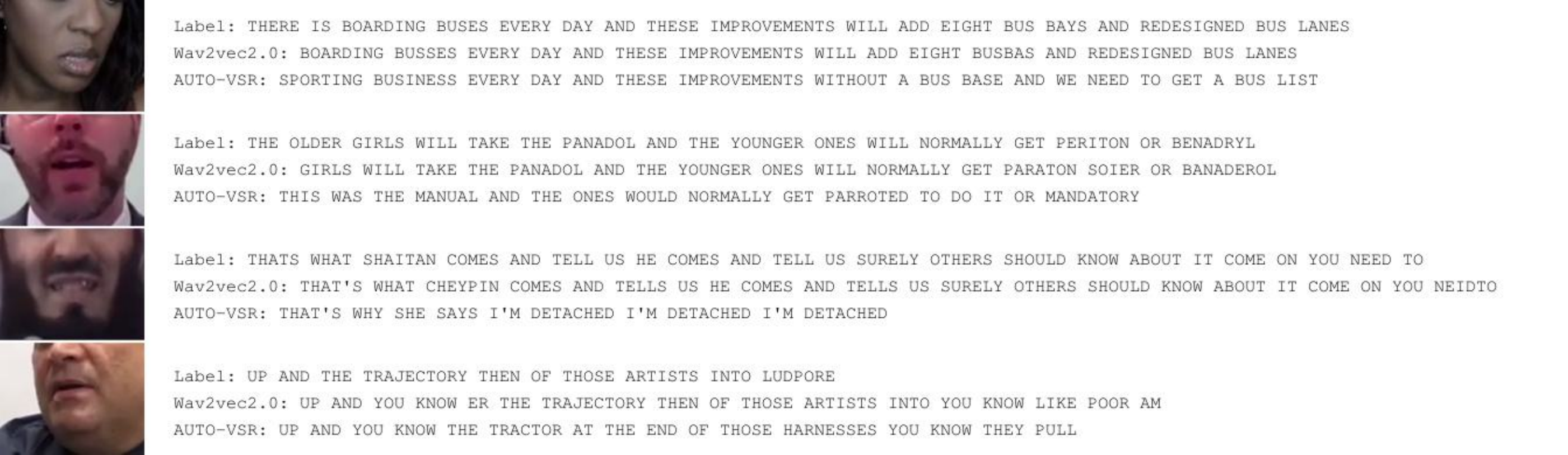}
    \caption{\textbf{Qualitative results.} The predictions of the Wav2vec2.0 and Auto-AVSR models on sample sequences from our test set. Wav2Vec2.0 mostly makes errors in terms of near-by homophones, \eg, PARATON  \vs PERITON in $2^{nd}$ row, LIKE POOR \vs LUDPORE in $4^{th}$ row. In comparison, the state-of-the-art VSR framework Auto-AVSR predictions deviate significantly from the target speech, \eg, SPORTING BUSINESS \vs THERE IS BOARDING BUSES in $1^{st}$ row.}
    \label{fig:results}
\end{figure*}

\section{Additional Results\label{sec:add_results}}

\begin{figure*}
    \centering
    \includegraphics[width=1.00\linewidth]{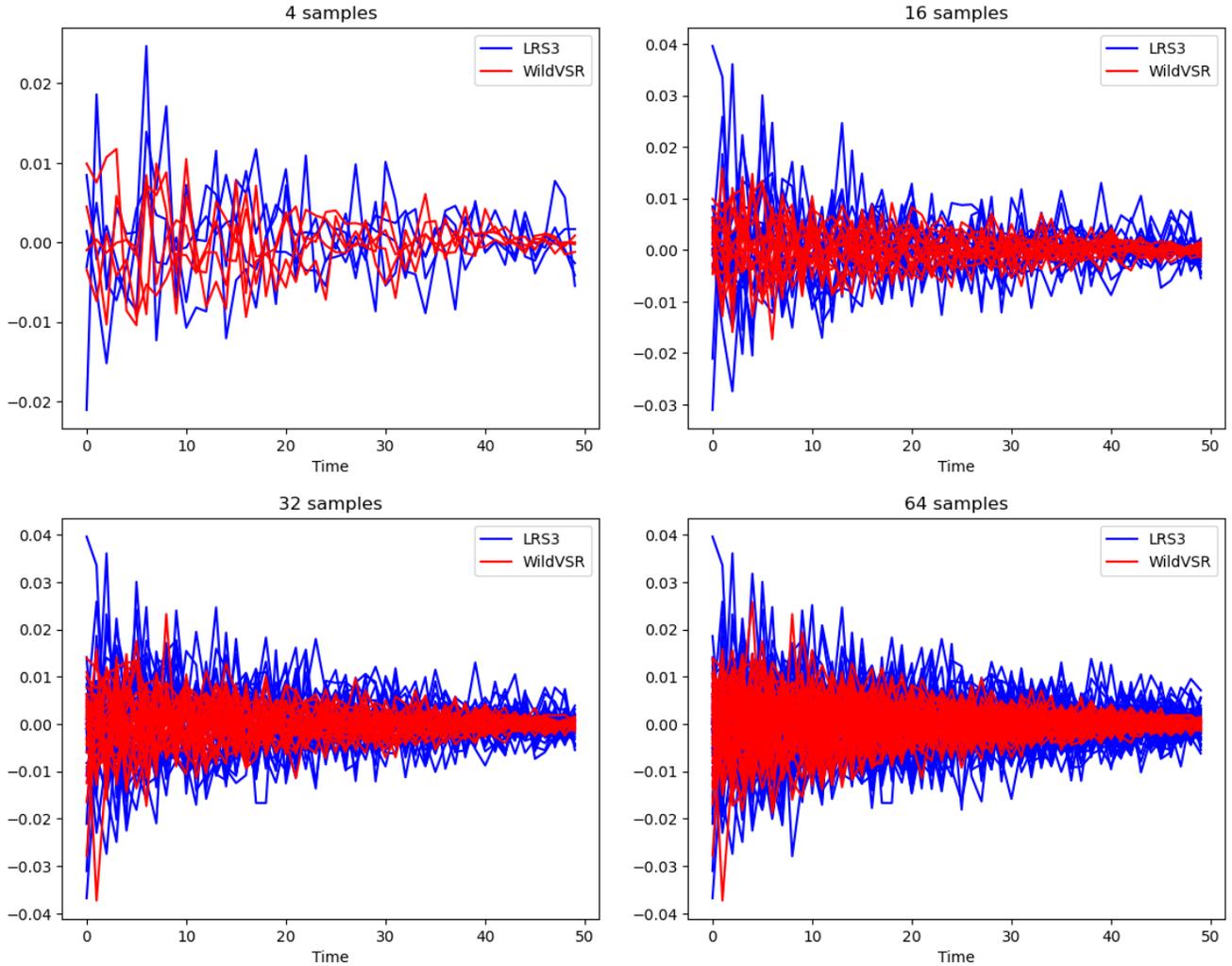}
    \caption{\textbf{Visualization of the dominant spatial mode of the tucker decomposition over time} of Auto-AVSR encoder representations on LRS3 (in blue) and WildVSR (in red) test sets. It can be seen that when adding more samples, the LRS3 representations envelop the WildVSR representations indicating that the salient modes of LRS3 are more compared to WildVSR.}
    \label{fig:tucker_2}
\end{figure*}

\noindent\textbf{ASR models:} We benchmark the Wav2Vec2.0 \cite{baevski2020wav2vec} and Whisper \cite{radford2022robust} on both LRS3 and our test sets. Wav2vec2.0 achieved $6.2$ and $23.4$ WER scores on both datasets respectively. In comparison, Whisper exhibits even higher performance with scores less than $4$ WER on both test sets. These low WER scores signify the models' ability to accurately transcribe speech, capturing the spoken words with remarkable precision. Moreover, the small standard deviation observed for both models suggests that their performance is consistently reliable, with minimal variation in recognition errors across different samples. However, Wav2vec2.0 follows the same trend as VSR models on our test set, deviating by 12 WER points from the LRS3 score. This discrepancy while not being attributed to visual features responsible for VSR models performance, is more likely to be influenced by the sequence of phonemes in our test set. It is possible that the transcriptions in our test set contain more challenging or complex sequences of phonemes, which may pose difficulties for the VSR models and result in a drop in their performance. Unlike Whisper, Wav2vec2.0 relies solely on character-level CTC decoding without the use of any language model to ensure valid word predictions. This lack of language modeling support in Wav2vec2.0 could contribute to its higher WER on our test set. Projecting on VSR approaches, we hypothesize that factors beyond visual features alone, such as the sequence of phonemes in our test set are also likely to contribute to the drop in performance.

\subsection{WildVSR-Word} 
To transform our sentence-level test set into a word-level format akin to the LRW (Lip Reading in the Wild) dataset \cite{chung2017lip}, we followed a systematic process. The objective was to ensure that the selected word segments were not only contextually relevant but also well-aligned with the LRW classes.

\begin{itemize}
    \item Whisper~\cite{radford2022robust} Word Boundaries: The primary challenge in transitioning from sentence to word level lies in identifying accurate word boundaries within continuous spoken sentences. To address this, we made use of whisper word boundaries. These boundaries provided a reliable temporal localization of individual words within the sentences, allowing us to select the start and end times of each word.

    \item LRW Class Overlap: Given that our aim is to align with the LRW word classes, we performed a filtering operation on the identified word boundaries. Only the words which  overlap with the LRW class vocabulary were retained for the next steps. This ensured that our WILDVSR-Word dataset is directly comparable and compatible with existing LRW models.

    \item Central Frame Extraction: To maintain consistency and ensure the best representation of each word, we centered the segment on the midpoint timestamp of each selected word boundary. From this center point, we cropped video segments to obtain a fixed length of 29 frames. This length was chosen to comply with the LRW creation process.

\end{itemize}

As shown in \reftab{word_level}, we tested models from \cite{ma2022training} on the resulting dataset, termed "WILDVSR-Word". In fact, we observe a similar drop in accuracy as in sentence-level VSR. The DCTCN drops by $60.0$ points, this confirms the generalization issues of VSR models for both sentence and word level.

\begin{table}[t]

\caption{\textbf{Performance comparison on word-level VSR.}}
\centering
\setlength{\tabcolsep}{8pt}
     \adjustbox{width=0.9\columnwidth}{

\begin{tabular}{l c | c}

\toprule
   \multirow{2}{*}{\textbf{Model}} & \multicolumn{2}{c}{\textbf{Test sets}} \\
    % \cline{2-3}
     & LRW & WildVSR-W  \\
\midrule
 DCTCN/Boundary \cite{ma2022training} & 92.1 & 34.6  \\
 DCTCN \cite{ma2022training} & 89.6 &   32.5 \\
 MSTCN \cite{ma2022training} & 88.9&   29.6 \\

\bottomrule
% \vspace{-0.5cm}
\end{tabular}
}
\label{word_level}
\end{table}

\begin{figure}
    \centering
    \includegraphics[width=1.00\linewidth]{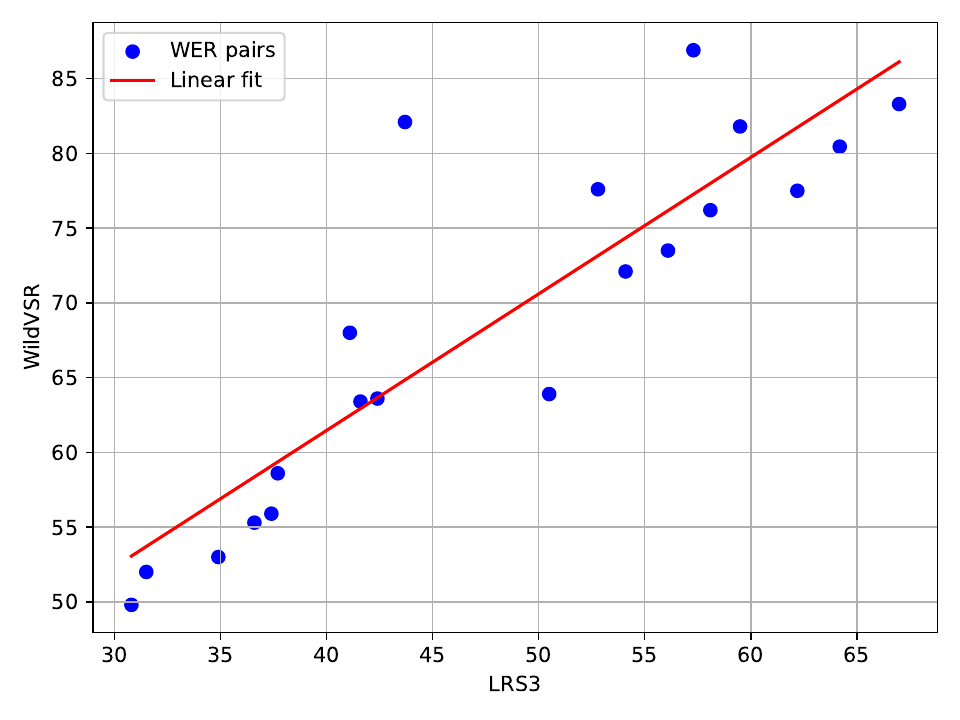}
    \caption{\textbf{Model performance on LRS3 \vs WildVSR}. Each data point corresponds to one model in ones we used in the main results. The plots
reveal two main phenomena: (i) There is a significant drop in accuracy from LRS3 to WildVSR. (ii) The model WERs closely follow a linear function with slope greater than 1 (1.30). This means that every WER decrease on LRS3  translates into more than one WER point on the new WildVSR test set.}
    \label{fig:linear}
\end{figure}

\section{Additional Details on FLOPs Computation\label{sec:training_compute}}
Here, we detail the approach employed for calculating the training budget (FLOPs) of the VSR/AVSR models in Table.~2 of the main paper. As discussed in Sec.~4 of the paper, we utilize the methodology described in~\cite{kaplan2020scaling} for estimating the compute budget. Accordingly, a transformer model's training compute for a single input token is approximated to be $6 N$, where $N$ denotes the number of model parameters. Briefly, it takes around $2N$ compute per token for the forward pass (the backward pass is approximately twice the compute as the forwards pass), resulting in a total of $6N$ compute per token for a single forward-backward computation. Consequently, the total training compute required is $C = 6 N \times D$, where $D$ denotes the total number of tokens the model is trained on. For the task of visual speech recognition, tokens refer to the number of input frames. Next, we describe the calculations for different models reported in Table.~2 of the main paper.

\subsection{Fully-supervised Models}
\noindent\textbf{Ma \etal~\cite{ma2022visual}:} This model has a total of $52.5$M parameters and is trained on $1459$ hours of video data for $50$ epochs. The $1459$ hours correspond to $131.3$M frames (\ie, $1459 \times 3600$ seconds $\times 25$ fps). Thus, the total compute required for $50$ epochs is $6 \times 52.5$M $\times (131.3\text{M}\times 50)$, which results in $2.1\times 10^{18}$ FLOPs, \ie,  $2.1$ exaFLOPs.

\noindent\textbf{Auto-AVSR~\cite{ma2023auto}:} This model has $250.1$M parameters and is trained for $75$ epochs. There are three variants of the model trained with different amount of data: $661$, $1759$ and $3448$ hours, corresponding to $59.5$M, $158.3$M and $310.3$M frames, respectively. As a result, the total training compute requirement for these $661$, $1759$ and $3448$ hours variants comes out as $6.7$, $17.8$ and $34.9$ exaFLOPs, respectively. 

\begin{table*}[t] 
    \centering
    \setlength{\tabcolsep}{6pt}

    \caption{\textbf{Training budget computation for RAVen~\cite{haliassos2022jointly} model variants.} `ST' refers to self-training, where finetuning is performed on $1759$ hours of labeled and pseudo-labeled data. The different data regimes of $30$, $433$ and $1759$ hours translate to $2.7$M, $39$M and $158.3$M frames, respectively. The encoder and decoder sizes are denoted by $N_e$ and $N_d$. Note that encoder size is doubled for pretraining ($2N_e$) due to the training of both audio and video encoders, while finetuning is with single encoder and decoder ($N_e + N_d$). Also see text in Section~\ref{sec:training_compute} for more details.\vspace{-0.2cm}}

     \adjustbox{width=\textwidth}{
    \begin{tabular}
    {l c c c  c | c c  c} 
     \toprule[0.1em]
    \multirow{2}{*}{\textbf{Model}} & \textbf{Encoder Size} &  \textbf{Pretraining} & \textbf{Decoder Size} & \textbf{Finetuning}  & \multicolumn{3}{c}{\textbf{Compute} (exaFLOPS)} \\
    & (M) &  epochs $\times$ $\#$ Frames (M) & (M) & epochs $\times$ $\#$ Frames (M) & \textbf{Pretraining} & \textbf{Finetuning } & \textbf{Total} \\
    \midrule
    & $N_e$ & $D_p$ & $N_d$ & $D_f$ & $C_p = 6\cdot 2 N_e D_p$ & $C_f = 6 (N_e + N_d) D_f$ & $C = C_p + Cf$ \\
    \midrule
    % \multicolumn{8}{c}{\myccnew AVHuBERT~\cite{shi2022learning} \textit{Low-resource Setting}} \\

    \multicolumn{8}{c}{\myccnew \textit{Low-resource Setting}} \\
    Base 433h & 52.4 &  150 $\times$ 39 & 10.1 & 50 $\times$ 2.7 & 3.6 & 0.05 & 3.7 \\
    Base 1759h  & 52.4 &  150 $\times$ 158.3 & 10.1 & 50 $\times$ 2.7 & 14.9 & 0.05 & 14.9 \\
    Large 1759h  & 339.3 &  150 $\times$ 158.3 & 10.2 & 50 $\times$ 2.7 & 96.6 & 0.2 & 96.8 \\
    Large 1759h (w/ ST)  & 339.3 &  150 $\times$ 158.3 & 153.3 & 50 $\times$ 158.3 & 96.6 & 35.1 & 131.7 \\
    \midrule
    \multicolumn{8}{c}{\myccnew \textit{High-resource Setting}} \\
    
    Base 433h  & 52.4 &  150 $\times$ 39 & 26.3 & 75 $\times$ 39 & 3.6 & 1.4 & 5.0 \\
    Base 1759h  & 52.4 &  150 $\times$ 158.3 & 26.3 & 75 $\times$ 39 & 14.9 & 1.4 & 16.3 \\
    Large 1759h  & 339.3 &  150 $\times$ 158.3 & 153.3 & 75 $\times$ 39 & 96.6 & 8.7 & 105.3 \\
    Large 1759h (w/ ST)  & 339.3 &  150 $\times$ 158.3 & 153.3 & 75 $\times$ 158.3 & 96.6 & 35.1 & 131.7 \\
     \bottomrule[0.1em]
    \end{tabular}}
    \label{training_compute}
\end{table*}

\begin{table*}[t] 

    \centering
    \setlength{\tabcolsep}{12pt}
    
    \caption{\textbf{Performance comparison on our proposed test set with varying the attributes}. We report the best variant per model.\vspace{-0.2cm}}

     \adjustbox{width=\textwidth}{
    
    \begin{tabular}
    { l | c c | c c | c c c | c c c} 
     \toprule[0.1em]
      \textbf{Method} &  \multicolumn{2}{c|}{\textbf{Accent}} & \multicolumn{2}{c|}{\textbf{Gender}} & \multicolumn{3}{c|}{\textbf{Age}} & \multicolumn{3}{c}{\textbf{Ethnicity}}\\
     & Native & Non-Native  & Male & Female & Young & Adult & Old & White & Black & Others \\

    \midrule
    
    Auto-AVSR~\cite{ma2023auto} &                       35.1 & 46.9 & 38.3 & 37.4 & 45.2 & 38.1 & 38.5 & 38.2 & 42.1 & 38.1\\
    AV-HuBERT~\cite{shi2022learning,shi2022robust} & 47.5 & 55.3 & 49.2 & 47.7 & 52.0 & 48.4 & 48.7 & 48.5 & 50.0 & 48.4  \\
    RAVen~\cite{haliassos2022jointly}  & 45.2 & 55.0 & 47.5 & 45.3 & 54.5 & 46.4 & 47.3 & 46.2 & 48.7 & 47.3 \\
    
     \bottomrule[0.1em]
    \end{tabular}

    }
    \label{attributes_ap}
\end{table*}

\subsection{SSL Pretrained and Finetuned Models}
\subsubsection{AV-HuBERT~\cite{shi2022learning}} 
The AV-HuBERT model has multiple variants corresponding to different model sizes and training data duration. The Base model has encoder and decoder with $103.3$M and $57.3$M parameters, while the Large model has $325.4$M and $151.9$M parameters, respectively. 

The Base model is pretrained for $5$ iterations for $0.4$M steps on $32$K frame tokens per step ($32$ GPUs with $1$K frame tokens per GPU). Differently, the Large model is initialized from the Base model (after $4$ iterations) and further pretrained for $1$ iteration for $0.6$M steps on $64$K frame tokens per step ($64$ GPUs with $1$K frame tokens per GPU). Consequently, pretraining the Base model for one iteration involves $6 \times 103.3\text{M} \times (0.4\text{M} \times 32\text{K})$ FLOPs, equal to $7.9$ exaFLOPs. Similarly, pretraining the Large model for one iteration involves $6 \times 325.4\text{M} \times (0.6\text{M} \times 64\text{K})$ FLOPs, equal to $74.9$ exaFLOPs. As a result, Base model pretraining on $5$ iterations takes $39.6$ exaFLOPs, while the Large model pretraining ($4$ base iterations followed by $1$ large iteration) requires $106.6$ exaFLOPs.

During finetuning in low-resource setting ($30$ hours), only the decoder is trained for $18$K steps with $8$K frame tokens per step ($8$ GPUs at $1$K frames per GPU). Thus, finetuning the Base model in low-resource setting takes $6 \times 57.3\text{M} \times (18\text{K} \times 8\text{K})$ FLOPs, equivalent to $0.05$ exaFLOPs. Similarly, finetuning the Large model takes $6 \times 151.9\text{M} \times (18\text{K} \times 8\text{K})$, resulting in $0.13$ exaFLOPs. As a result, combining both pretraining and finetuning compute requirements, in the low-resource setting, Base and Large models require $39.7$ and $106.7$ exaFLOPs, respectively.

In contrast, in the high-resource setting ($433$ hours), the encoder is trained for $22.5$K steps while the decoder is trained for $45$K steps with $8$K frame tokens per step. Thus, finetuning the Base model in high-resource setting needs $6 \times [(103.3\text{M} \times (22.5\text{K} \times 8\text{K}) +  57.3\text{M} \times (45\text{K} \times 8\text{K})]$ FLOPs, equivalent to $0.23$ exaFLOPs.
Similarly, finetuning the Large model takes $6 \times [(325.4\text{M} \times (22.5\text{K} \times 8\text{K}) +  151.9\text{M} \times (45\text{K} \times 8\text{K})]$, equal to $0.7$ exaFLOPs. Consequently, adding both pretraining and finetuning compute requirements, in the high-resource setting, Base and Large models require $39.9$ and $107.3$ exaFLOPs, respectively.

\subsubsection{RAVen~\cite{haliassos2022jointly}} The RAVen model has Base and Large variants trained on different data regimes in the pretraining and finetuning stages. The Base model for low-resource setting has $52.4$M and $10.1$M parameters in the encoder and decoder. For Base model in high-resource, the encoder is same while the decoder is larger at $26.3$M parameters. Similarly, the Large model in low-resource setting has $339.3$M and $10.2$M parameters for encoder and decoder. While the Large model in high-resource setting has $339.3$M and $153.3$M parameters for encoder and decoder. The self-trained variant has same sizes as Large variant in high-resource setting. Also, it is important to note that the pretraining involves training the audio and video encoders together (\ie, twice the encoder parameters $2N_e$) and finetuning utilizes a single encoder and decoder ($N_e + N_d$). Furthermore, while $150$ epochs are used during pretraining for all models, the low-resource and high-resource models are finetuned for $50$ and $75$ epochs, respectively. The different data regimes of $30$, $433$ and $1759$ hours translate to $2.7$M, $39$M and $158.3$M frames, respectively. 

Table~\ref{training_compute} reports the compute required for different variants of the RAVen model. The Base model pretrained on $433$ hours in low-resource setting ($30$ hour finetuning) utilizes $6 \times (2\times 52.4\text{M}) \times 39\text{M} \times 150 $ FLOPs for pretraining and $6 \times (52.4 + 10.1)\text{M} \times 2.7\text{M} \times 50$ FLOPs for finetuning, resulting in $3.7$ exaFLOPs in total. Similarly, the Large model pretrained on $1759$ hours and finetuned on $433$ hours (high-resource) requires $6 \times (2\times 339.3\text{M}) \times (158.3\text{M} \times 150) $ FLOPs for pretraining and $6 \times (339.3 + 153.3)\text{M} \times (39\text{M} \times 75)$ FLOPs for finetuning, \ie, a total of $105.3$ exaFLOPs. Furthermore, since self-training involves pseudo-labeling the unlabeled data and utilizing them during finetuning, all $1759$ hours are used for finetuning. Consequently, self-trained Large models require $35.1$ exaFLOPs during finetuning, resulting in $126.1$ exaFLOPs requirement for the entire training.

\end{document}